\documentclass[final,2p,times]{elsarticle}

\usepackage{amssymb}
\usepackage{fullpage} 

\usepackage[toc,page]{appendix}
\appto\appendix{\addtocontents{toc}{\protect\setcounter{tocdepth}{1}}}

\usepackage{amsmath,amssymb,amsxtra,mathtools,amsthm}
\usepackage{array}
\usepackage{graphicx}
\usepackage{algorithm}
\usepackage{algpseudocode}
\usepackage{dsfont}

\usepackage{verbatim}
\usepackage{enumerate}
\usepackage{parskip}

\usepackage{color}
\definecolor{clemson-orange}{RGB}{234,106,32}
\definecolor{chicago-maroon}{RGB}{128,0,0}
\definecolor{cincinnati-red}{RGB}{190,0,0}
\definecolor{soft-cyan}{RGB}{68,85,90}
\definecolor{firebrick}{RGB}{178,34,34}
\definecolor{crimson}{RGB}{220,20,60}

\usepackage{hyperref}
\hypersetup{
  colorlinks   = true,    
  urlcolor     = blue,    
  linkcolor    = firebrick,    
  citecolor    = blue      
}

\usepackage{parskip}

\newtheorem{theorem}{Theorem}[section]
\newtheorem{lemma}[theorem]{Lemma}
\newtheorem{corollary}[theorem]{Corollary}

\newtheorem{definition}{Definition}
\newtheorem{remark}{Remark}

\newif\ifsolutions \solutionstrue

\newcommand{\poly}{\textrm{poly}}

\newcommand{\abs}[2][]{#1\lvert#2 #1\rvert}
\newcommand{\norm}[2][]{#1\lVert #2 #1\rVert}
\newcommand{\tuple}[2][]{#1 \langle #2 #1 \rangle}
\newcommand{\ip}[3][]{\tuple[#1]{#2,#3}}
\DeclareMathOperator*{\relu}{{\textrm{ReLU}}}

\def\ve#1{\mathchoice{\mbox{\boldmath$\displaystyle\bf#1$}}
{\mbox{\boldmath$\textstyle\bf#1$}}
{\mbox{\boldmath$\scriptstyle\bf#1$}}
{\mbox{\boldmath$\scriptscriptstyle\bf#1$}}}

\newcommand{\x}{{\ve x}}
\newcommand{\y}{{\ve y}}
\newcommand{\z}{{\ve z}}
\renewcommand{\v}{{\ve v}}
\newcommand{\g}{{\ve g}}

\renewcommand{\c}{{\ve c}}

\newcommand{\w}{{\ve w}}

\newcommand{\s}{\textrm{s}}
\newcommand{\E}{\mathbb{E}}

\newcommand{\bb}{\mathbb}
\newcommand{\R}{\bb R}

\newcommand{\ind}[1]{\mathds{1}_{\left\{#1\right\}}}

\newcommand{\ignore}[1]{}

\begin{document}

\begin{frontmatter}



\title{Provable Training of a $\relu$ Gate with an Iterative Non-Gradient Algorithm}




\author[inst1]{Sayar Karmakar}

\affiliation[inst1]{organization={Department of Statistics, University of Florida},
            addressline={230 Newell Drive}, 
            city={Gainesville},
            postcode={32611}, 
            state={FL},
            country={USA}}

\author[inst2]{Anirbit Mukherjee \fnref{Hello}}

\affiliation[inst2]{organization={Department of Computer Science, The University of Manchester},
            addressline={Kilburn Building}, 
            city={Manchester},
            postcode={M13 9PL}, 
            country={U.K.}}

\fntext[Hello!]{Most of this work was done when Anirbit was at Wharton, the Department of Statistics at UPenn and at the Department of Applied Mathematics and Statistics, J.H.U.}

\begin{abstract}
 In this work, we demonstrate provable guarantees on the training of a single $\relu$ gate in hitherto unexplored regimes. We give a simple iterative stochastic algorithm that can train a $\relu$ gate in the realizable setting in {\it linear time}  while using significantly milder conditions on the data distribution than previous such results. 
 
 
  Leveraging certain additional moment assumptions, we also show a first-of-its-kind approximate recovery of the true label generating parameters under an (online) data-poisoning attack on the true labels, while training a $\relu$ gate by the same algorithm.  Our guarantee is shown to be nearly optimal in the worst case and its accuracy of recovering the true weight degrades gracefully with increasing probability of attack and its magnitude. 
  
  For both the realizable and the non-realizable cases as outlined above, our analysis allows for mini-batching and computes how the convergence time scales with the mini-batch size. We corroborate our theorems with simulation results which also bring to light a striking similarity in trajectories between our algorithm and the popular S.G.D. algorithm - for which similar guarantees as here are still unknown.


\end{abstract}


\end{frontmatter}


\clearpage 

{\small 
\tableofcontents 
}

\section{Introduction}
\label{intro}

Over the last few years, there has been a surge of activity in using neural networks for complex artificial intelligence tasks. Human world champions of classic hard board games have famously been defeated by neural net-based approaches, the \cite{silver2016mastering, silver2017mastering,silver2018general,schrittwieser2020mastering}. At the core of many of these successes lie the ability of various heuristics to be able to solve the \textit{learning theory} question of function optimization/\textit{risk minimization}, 
\begin{align}
\min_{\mathbf{N} \in {\cal N}} \mathbb{E}_{\z \in {\cal D}} [\ell (\mathbf{N}, \z)]
\end{align}
where $\ell$ is some lower-bounded non-negative function, members of ${\cal N}$ are continuous piecewise linear functions representable by some chosen neural net architecture and we only have sample access to the distribution ${\cal D}$. This reduces to the \textit{empirical risk minimization} question when this ${\cal D}$ is a uniform distribution on a finite set of points. But as of today, we have little or no mathematical guarantees about these heuristics which seemingly very efficiently solve the many useful instances of these optimization problems. 

To the best of our knowledge about the state-of-the-art in deep-learning theory, any of these two optimization problems is typically {\it provably} solvable in poly-time for nets with more than $1$ neuron in either of the following two mutually exclusive scenarios : {\bf (a)} the nets in the class ${\cal N}$ are of constant size and the data comes as tuples $\z = (\x,\y)$ with $\y$ being the noise corrupted output at input $\x$ for a net (of a known architecture that which would be common to the class ${\cal N}$). And {\bf (b)} the nets in ${\cal N}$ would be asymptotically large and the data comes as tuples $\z = (\x,\y)$ with no explicit functional relationship between $\x$ and $\y$ (but there could be geometric or statistical assumptions about the $\x$ and $\y$).

 The simplifications that happen for infinitely large networks have been discussed since \cite{neal1996priors} and this theme has had a recent resurgence in works like \cite{chizat2018global,jacot2018neural}. Eventually this led to an explosion of literature in getting linear time training of various kinds of neural nets when their width is a high degree polynomial in training set size and inverse accuracy (a somewhat {\it unrealistic} regime),  \cite{lee2017deep,wu2019global,du2018gradient,su2019learning,kawaguchi2019gradient,huang2019dynamics,allen2019convergenceDNN,allen2019learning,allen2019convergenceRNN,du2018power,zou2018stochastic,zou2019improved,arora2019exact,arora2019harnessing,li2019enhanced,arora2019fine}. The essential proximity of this regime to kernel methods have been thought of separately in works like  \cite{allen2019can,wei2019regularization}. 
 
 On the other hand we note that in the fully agnostic setting training even a single $\relu$ gate can be SPN-hard as shown in \cite{goel2016reliably}. Hence its an interesting mathematical question to isolate general conditions when the convergence speed can be fast for a single $\relu$ gate.



To the best of our knowledge, for training a single neuron to $\epsilon-$accuracy, existing results until this work were restricted to a sample complexity of ${\cal O}(\poly(1/\epsilon))$ for (Stochastic) Gradient Descent ((S.)G.D.) even with realizable data. And any improvements to this had been known to happen only for the case of the marginal distribution on the input being Gaussian. We refer the interested readers to \cite{frei2020agnostic} for a comprehensive summary of these results - against many of which we will compare our results too. In this paper, we break this barrier and improve the sample-complexity of training a single $\relu$ gate to ${\cal O}(\log(1/\epsilon))$ for realizable data and {\it without} tying ourselves to any specific distribution. We emphasize that not only are we able to achieve this only by slightly tweaking the popular S.G.D. algorithm itself but also that our algorithm has guarantees in cases where we make the data non-realizable by allowing for a data-poisoning attack. Our distributional assumptions are mild and reminiscent of the subspace eigenvalue conditions from \cite{dulitian}. Moreover, through thorough experiments, we will show that our modified S.G.D. has strikingly similar convergence features as the traditional S.G.D. We summarize the technical details of our results in the following subsection.

\subsection{A summary of our results}\label{sec:summary}

To make progress with provable training of a single gate we draw inspiration from the different avatars of iterative stochastic non-gradient algorithms analyzed in the past,  \cite{rosenblatt1958perceptron,Pal1992MultilayerPF,freund1999large,kakade2011efficient,klivans2017learning,goel2017learning,goel2018learning}.   

We shall organize our contributions in this paper under four groups as follows :  
\\

{\it Firstly,} in the short Section \ref{sec:glmtron} we start with a quick re-analysis of a known algorithm called the {\rm GLM-Tron}  \cite{kakade2011efficient} but under more general conditions than the previous proofs about it. We show how well it can do (empirical) risk minimization on any Lipschitz gate with Lipschitz constant $<2$ in the noisily realizable setting while no assumptions are being made on the distribution of the noise beyond their boundedness - hence the noise can be {\textit{adversarial}}. We also point out how the result can be improved under certain benign assumptions on the noise. 


{\it Secondly,} in Section \ref{sec:almostSGDReLU}, we exclusively focus on training the $\relu$ gate, $\R^n \ni \x \mapsto \max \{0,\w^\top \x \} \in \R$ for $\w \in \R^n$ being its weight. We note that for this gate, the corresponding empirical or the population risk is neither convex nor smooth w.r.t. how it depends on the weights. And yet we show a very simple iterative stochastic algorithm which can provably recover in linear time the underlying parameter $\w_*$ of the $\relu$ gate when the data being sampled is exactly realizable of the form $(\x, \max \{ 0, \w_*^\top\x \} )$. That is w.h.p. in $\log \left (\frac{1}{\epsilon} \right)$ iterations we get $\epsilon$ close to $\w_*$ while starting from any arbitrary initial point. (We recall that linear time convergence i.e getting $\epsilon$ close to the global minima in ${\cal O} (\log(\frac{1}{\epsilon}))$ time is a hallmark of specialized optimization methods adapted for smooth strongly convex objectives like \cite{johnson2013accelerating}). To achieve this we use a mild distributional condition which essentially captures the intuition that enough of our samples are such that $\w_*^\top\x >0$. To the best of our knowledge, this is the first example of nearly distribution-free training of a $\relu$ gate in linear time.

Note that, in Section \ref{sec:almostSGDReLU} we are using a stochastic algorithm while solving a regression problem specific to a $\relu$ gate and are exploiting the structure of the $\relu$ gate (and mild distributional assumptions) to directly achieve parameter recovery. The results in Section \ref{sec:glmtron} also apply to a $\relu$ gate as a special case but in contrast, therein we used full-batch iterative updates to gain other advantages, namely of being able to handle more general gates while having essentially no distributional assumptions on the training data.

 {\it Thirdly,} by making a slightly stronger distributional assumption, in Case (II) of the Theorem \ref{thm:dadushrelu:noise} in Section \ref{sec:almostSGDReLU} we also encompass the case when during training the oracle behaves adversarially i.e it tosses a biased coin and decides whether or not to additively distort the true labels by a bounded perturbation. Additionally, we also allow for the bias of the adversary's coin to be data-dependent. This is a ``data-poisoning" attack since the adversary corrupts the training data in an online fashion. In this case, we show that the accuracy of the algorithm in recovering $\w_*$ is not only worst-case near optimal but is such that the accuracy degrades gracefully as the probability of the adversary's attack or the magnitude of the distortion increases. 
 
 To the best of our knowledge, this is the first guarantee on training a $\relu$ gate while under any kind of an adversarial attack. Also in both these cases above we allow for mini-batching in the algorithm and keep track of how the mini-batch size affects the convergence time. 
 
 {\it Lastly,} in Section \ref{sec:exp_relu} we give an experimental demonstration of the performance of our algorithm. We do a side-by-side comparison on a $\relu$ gate between S.G.D. and our modified S.G.D., under various setting which fall under the ambit of Theorem \ref{thm:dadushrelu:noise}. In particular we track how the distance to the original optima ($\w_*$) changes with time for the various settings that we consider. Seen from this perspective we emphasize that while guarantees like Case (II) of Theorem \ref{thm:dadushrelu:noise} still remain unknown for S.G.D., our algorithm's behaviour in experiments closely resembles that of  S.G.D. under similar settings. Thus our experiments encourage the conjecture that maybe our modification keeps unchanged the stochastic process induced by S.G.D. on a $\relu$ gate. We leave it for future work to investigate this possibility and to try generalizing this for larger nets.  

\subsection{Comparison to concurrent literature}

Firstly, we note that the result in \cite{goel2018learning} includes as a special case, learning a $\relu$ gate under realizable settings - but only under the assumption of the distribution being symmetric. Specific to the marginal distribution on the data being Gaussian, works like \cite{soltanolkotabi2017learning,kalan2019fitting} had solved the same problem using gradient-based methods. 

A notable recent progress with understanding the behaviour of (stochastic) gradient descent on a $\relu$ gate was achieved in \cite{frei2020agnostic}. Their Theorem D.1 (b) is solving a similar question as our Theorem \ref{thm:dadushrelu:noise} Case (I). But our algorithm, in this special case, not only accounts for the effect of mini-batching on the convergence time but also converges exponentially faster than what is guaranteed in \cite{frei2020agnostic}.

Also significantly in contrast to these previous results cited above, our Theorem \ref{thm:dadushrelu:noise} Case (II) encompasses the situation of a probabilistic adversary causing distortions to the true labels. To the best of our knowledge this is the first work to analyze training of a $\relu$ gate in any kind of adversarial setup - in particular a data-poisoning attack on the training data (labels). We also allow for the adversary to decide to attack or not using a biased coin toss whose bias is allowed to be data-dependent. 

Lastly, unlike any of these previous results, we keep track of the subtleties of using mini-batches and how the mini-batch size affects the convergence time.

In  \cite{diakonikolas2020approximation}, the authors had given algorithms for learning of a $\relu$ gate in the non-realizable setting for certain nice marginal distributions on the data. We note that such results about risk minimization are incomparable to our goal in Theorem \ref{thm:dadushrelu:noise} Case (II) of recovering the generating weights (the $\w_*$ therein) as closely as possible under adversarial corruption of the training labels. But this result of ours can be seen as a natural regression analogue of the recent result in \cite{diakonikolas2020learning}  about learning half-space indicators under a Massart noise.

\section{Re-analyzing the GLM-Tron}\label{sec:glmtron}

In this section we shall take a relook at the GLM-Tron algorithm (given below) from \cite{kakade2011efficient} and show that it converges on certain Lipschitz gates with no distributional assumption on the data.

\begin{algorithm}[H]
\caption{GLM-Tron}
\label{GLMTron} 
\begin{algorithmic}[1]
\State {\bf Input:} {$\{ (\x_i,y_i) \}_{i=1,\ldots,m}$ and an \textit{activation function}  $\sigma : \R \rightarrow \R$}
\State {$\w_1 =0$}
\For{$t = 1,\ldots$}
\State {$\w_{t+1} := \w_t + \frac {1}{m} \sum_{i=1}^m \Big( y_i - \sigma(\langle \w_t , \x_i\rangle) \Big)\x_i$}
\EndFor
\end{algorithmic}
\end{algorithm}

First, we state the following crucial lemma, 

\begin{lemma}\label{stepdecrease} 
Assume that for all $i = 1,\ldots,S$ $\norm{\x_i} \leq 1$ and in Algorithm \ref{GLMTron}, $\sigma$ is a $L-$Lipschitz non-decreasing function. Suppose the vector $\w$ and the scalar $W$ are s.t at iteration $t$, we have $\norm{\w_t - \w} \leq W$ and we define $\eta >0$ s.t $\norm{\frac {1}{S} \sum_{i=1}^S \Big ( y_i - \sigma (\langle \w, \x_i \rangle) \Big ) \x_i} \leq \eta $. Then it follows that, 

\[  \norm{\w_{t+1} - \w}^2 \leq \norm{\w_t - \w}^2 -  \Big ( \frac {2}{L} - 1 \Big ) \tilde{L}_S(h_t) + \Big (\eta^2 + 2\eta W(L + 1) \Big ) \]

where we have defined,\\ $\tilde{L}_S(h_t) := \frac {1}{S} \sum_{i=1}^S \Big (h_t(x_i) - \sigma (\langle  \w , \x_i\rangle) \Big)^2 = \frac {1}{S} \sum_{i=1}^S \Big ( \sigma (\langle \w_t, x_i\rangle ) - \sigma (\langle  \w , \x_i\rangle) \Big)^2$
\end{lemma}
\bigskip 

We give the proof of the above lemma in Appendix \ref{app:glmtronstep}. The above Algorithm \ref{GLMTron}
was introduced in \cite{kakade2011efficient} for bounded activations. Here we show the applicability of that idea for more general activations and also while having adversarial attacks on the labels. We will see in the following theorem as to how the above lemma leads to convergence of the \textit{effective-E.R.M.}, $\tilde{L}_S$ by GLM-Tron on a single gate.

\begin{theorem}{\bf (GLM-Tron (Algorithm \ref{GLMTron}) solves the effective-E.R.M. on a ReLU gate up to noise bound with minimal distributional assumptions.)}\label{GLMTronReLU} 
Assume that for all $i = 1,\ldots,S$ $\norm{\x_i} \leq 1$ and the label of the $i^{th}$ data point $y_i$ is generated as, $y_i = \sigma (\langle \w_*, \x_i \rangle ) + \xi_i$ s.t $\forall i, \vert \xi_i \vert \leq \theta$ for some $\theta \geq 0$ and $\w_* \in \R^n$.  If $\sigma$ is a $L-$Lipschitz non-decreasing function for $L < 2$ then in at most $T = \frac {\norm{\w_*}}{\epsilon}$ GLM-Tron steps we would attain parameter value $\w_T$ s.t,  

\[ \tilde{L}_S(h_T) =  \frac {1}{S} \sum_{i=1}^S \Big ( \sigma(\langle \w_T, x_i\rangle ) - \sigma (\langle  \w_* , \x_i\rangle) \Big)^2 < \frac{L}{2-L} \Big ( \epsilon +  (\theta^2 + 2\theta \cdot \norm{\w_*} \cdot (L + 1) ) \Big ) \]
\end{theorem}


The proof of the above theorem is deferred to  Appendix \ref{app:GLMTronReLU}. 

\textbf{Remark:} {\it Firstly,} note that in the realizable setting i.e when $\theta =0$, the above theorem is giving an upperbound on the number of steps needed to solve the ERM on say a $\relu$ gate to $O(\epsilon)$ accuracy. {\it Secondly,} observe that the above theorem does not force any distributional assumption on the $\xi_i$ beyond the assumption of its boundedness. Thus the noise could as well have been chosen \textit{adversarially} up to the constraint on its norm.


If we make some assumptions on the noise being benign then we can get the following. 

\begin{theorem}{\bf (Performance guarantees on the GLM-Tron (Algorithm \ref{GLMTron}) when solving E.R.M.)}\label{GLMTronReLUNoise}
Assume that the noise random variables $\xi_i, i = 1,\ldots,S$ are identically distributed as a centered random variable say $\xi$. Then for $T = \frac{\norm{\w_*}}{\epsilon}$, we have the following guarantee for GLM-Tron on the empirical risk after $T$ iterations (say $L_S(h_T)$), 

\[ \mathbb{E}_{\{ (\x_i , \xi_i) \mid i=1,\ldots S \}} \Big [ L_S(h_T) \Big ] \leq \E_{\xi} [\xi^2] + \frac{L}{2-L} \Big ( \epsilon +  (\theta^2 + 2\theta \cdot \norm{\w_*} \cdot (L + 1) ) \Big ) \]  
\end{theorem}


The proof for the above has been given in Appendix \ref{app:GLMTronReLUNoise}. Here we note a slight generalization of the above that can be easily read off from the above.

\begin{corollary}
Suppose that the joint distribution of $\{ \xi_i \}_{i=1,\ldots,S}$ is s.t $\mathbb{P} \Big [ \vert \xi_i \vert \leq \theta ~\forall i \in \{1,\ldots,S \} \Big ] \geq 1 - \delta$ Then  the guarantee of the above Theorem \ref{GLMTronReLUNoise} still holds but now with probability at least $1-\delta$ over the noise distribution.  
\end{corollary}

\ignore{
In Section \ref{sec:NeuroTronD2FBSingle} we will see a significant generalization of the above to multi-gate settings with non-realizable data - but while making some mild symmetry assumptions about the data and constraining the activation $\sigma$ to be of the form of a {\rm Leaky-$\relu$}. {\it We will see that the multi-gate version of the above analysis leads to a more complicated dynamical system to be analyzed and hence to much richer insights into the behaviour of large neural nets.}
}

In the next section we shall continue with the current theme of training a single neuron and see how a stochastic algorithm can be designed to get stronger training guarantees specific to a $\relu$ gate.  

\section{Learning a ReLU gate in the realizable setting and under a data-poisoning attack}\label{sec:almostSGDReLU}

In this section we consider an adversary executing a data-poisoning attack on an iterative stochastic learning algorithm (Algorithm \ref{dadushrelu:noise}) . Given a marginal distribution ${\cal D}$ on the inputs $\x$, suppose the corresponding true labels are generated as $y = \relu(\w_*^{\top} \x)$ for some unknown $\w_* \in \R^n$.
We assume sampling access to $\mathcal{D}$ and an adversarial label oracle that on the $t^{th}-$iterate gets queried with $b$ inputs $\{\x_{t_1},\ldots,\x_{t_b} \}$ drawn uncorrelatedly from ${\cal D}$. The oracle then flips a coin  for each minibatch data point with probability of the coin returning $0$  being $1-\beta(\x_{t_i})$ for some fixed function  $\beta : \R^n \rightarrow [0,1]$. We assume that these coin flips are uncorrelated to each other and the mini-batch sample and if the coin flip gives $1$ only then does the adversary do a bounded (by a constant $\theta_*$) additive distortion to the true label of the corresponding data.  


To learn the true labeling function $\R^n \ni \y \mapsto \relu(\w_*^\top \y) \in \R$ in this adversarially corrupted realizable setting we try to solve the following optimization problem, 
$\min_{\w \in \R^n} \E_{\x \sim {\cal D}} \Big [ \Big ( \relu(\w^\top \x) - y \Big )^2\Big]$

In contrast to previous work, we show that the simple algorithm given below solves this learning problem by leveraging the intuition that if we see enough labels $y = \relu(\w_*^\top \x) + \xi$ where $y>\theta_*$, then solving the linear regression problem on this subset of samples, gives a $\tilde{\w}_{*}$ which is close to $\w_*$.
In the situation, with adversarial corruption ($\theta_* >0$) we show in subsection \ref{opt:dadush} that our recovery guarantee is optimal in a certain sense. Additionally in the realizable case ($\theta_* =0$ or $\beta =0$ identically), our setup learns to arbitrary accuracy the true weight $\w_*$ using much milder distributional constraints than previous such results that we are aware of.

\begin{algorithm}
	\caption{\newline Modified mini-batch SGD for training a $\relu$ gate with adversarially perturbed realizable labels.}
	\label{dadushrelu:noise}
	\begin{algorithmic}[1]
		\State {\bf Input:} Sampling access to a distribution ${\cal D}$ on $\R^n$ and a function $\beta : \R^n \rightarrow [0,1]$
		\State {\bf Input:} Oracle access to labels $y \in \R$ when queried with some $\x \in \R^n$  
		\State {\bf Input:} An arbitrarily chosen starting point of $\w_1 \in \R^n$ 
		\For {$t = 1,\ldots$}
		\State Sample independently $s_t \coloneqq \{\x_{t_1},\ldots,\x_{t_b} \} \sim {\cal D}$ and query the oracle with this set. 
		\State The Oracle samples $\forall i=1,\ldots,b, \alpha_{t_i} \sim \{0,1\}$ with probability $\{ 1 -\beta(\x_{t_i}), \beta(\x_{t_i})\}$
		\State The Oracle replies $\forall i=1,\ldots,b, y_{t_i} = \alpha_{t_i} \cdot \xi_{t_i} + \relu(\w_*^{\top}\x_{t_i})$ s.t $\vert \xi_{t_i} \vert \leq \theta_*$ 
		
		\State Form the gradient (proxy), 
		\[\g_t := - \frac{1}{b} \sum_{i=1}^b \ind{y_{t_i} > \theta_*} (y_{t_i} - \w_t^\top \x_{t_i})\x_{t_i}\]
		\State $\w_{t+1} := \w_t - \eta \g_t$
		\EndFor 
	\end{algorithmic}
\end{algorithm}

We note that the choice of $\g_t$ in Algorithm \ref{dadushrelu:noise} resembles the stochastic gradient that is commonly used and is known to have great empirical success. In a true S.G.D., the indicator occurring in $\g_t$ would have been $\ind{\w_t^\top \x_{t_i} >0}$ for each $i$

Towards stating our theorems we define the following notation. 

\begin{definition}\label{def:dadush} 
Given $\w_* \in \R^n, \theta_* \in \R^+$, a distribution ${\cal D}$ on $\R^n$ and a function $\beta : \R^n \rightarrow [0,1]$,  we define the following constants associated to them (assuming they are finite),
\[ a_i \coloneqq \E_{\x \sim {\cal D}} \Big [ {\bf 1}_{\w_*^\top \x > 0} \norm{\x}^i  \Big ], \text{ for }i = 2,4\]
\[ \beta_{j} \coloneqq \E_{\x \sim {\cal D}} \Big[\beta (\x)  {\bf 1}_{\w_*^\top \x > 0}  \norm{\x}^j  \Big ], \text{ for }j = 1,2,3  \] 
\[ \lambda_1 (\theta_*) \coloneqq  \lambda_{\min}\Bigg(\mathbb{E}_{\x \sim {\cal D}}\Big[ {\bf 1}_{\w_*^\top \x > 2 \theta_*} \x  \x^\top \Big]\Bigg) \]
\end{definition} 

\begin{theorem}{\bf (Training a $\relu$ gate with realizable data and a probabilistic data-poisoning adversary. (Proof in Appendix \ref{sec:appendixa}))}\label{thm:dadushrelu:noise}
~\\
In Algorithm \ref{dadushrelu:noise} we will assume that (a) for $i \neq j$ and for all $t$, the random variables/data samples $\x_{t_i}$ and  $\x_{t_j}$ are uncorrelated and (b) that the random variables $\alpha_{t_i}$ and $\alpha_{t_j}$ are mutually uncorrelated and also uncorrelated with the the mini-batch choice $s_t$.  

\textbf{Case I : Realizable setting, $\theta_*=0$}. 

Suppose (a) $\mathbb{E}\Big[  \norm{\x}^4 \Big ]$ and the covariance matrix $\E \Big [ \x  \x^\top \Big ]$ exist and (b) $\w_*$ is s.t  $a_4$ exists and $ \mathbb{E}\Big[ {\bf 1}_{\w_*^\top \x >0 } \x  \x^\top \Big]$ is positive definite - and hence $\lambda_1 \coloneqq \lambda_1(0)$ is well defined. Then if $\lambda_1<\infty$, one can find a suitable step-size $\eta>0$ and run {\bf Algorithm \ref{dadushrelu:noise}}  starting from arbitrary $\w_1 \in \R^n$ so that $\forall \epsilon >0$, $\delta \in (0,1)$, after ${\rm T} = O \Big ( \log \frac{\norm{\w_1 - \w_*}^2}{\epsilon^2 \delta}  \Big)$ iterations we have 
\[\mathbb{P} \Big [  \norm{\w_{\rm T} - \w_*}^2   \leq \epsilon^2 \Big ] \geq 1 - \delta \]


\clearpage 
\textbf{Case II : With bounded adversarial corruption of the true labels,  $\theta_* >0$}

Suppose $\w_*$ and $\theta_*$ are such that (a) $a_2,a_4,\beta_1 (>0),\beta_2,\beta_3$ exist and (b)  $\lambda_1(\theta_*) > 0 $. Then there exists constants $b_1',c_1',c_2',c_3'$ (to be defined below) s.t. one can choose $\eta=\frac{b_1'}{\gamma c_1'}$ and run {\bf Algorithm \ref{dadushrelu:noise}}  starting from arbitrary $\w_1 \in \R^n$ so that, after ${\rm T} = O \left( \log \frac{\norm{\w_1 - \w_*}^2}{\epsilon^2 \delta- \theta_*^2 \cdot \Big(\frac{\frac{c_2'}{c_1'}+\gamma \cdot  \frac{c_3'}{b_1'}}{\gamma - 1}\Big)}  \right)$ iterations we have 
\[\mathbb{P} \Big [  \norm{\w_{\rm T} - \w_*}^2   \leq \epsilon^2 \Big ] \geq 1 - \delta \]

where  $\epsilon >0$ and $\delta \in (0,1)$ are s.t. 

\begin{align}
\epsilon^2 \delta = \beta_1^2 \cdot \frac{K \cdot \theta_*^2}{(2 \lambda_1(\theta_*)- \frac{1}{K})}
\end{align}

and $K>0$ large enough s.t $2\lambda_1(\theta_*) > \frac{1}{K}$,  and 

\begin{gather}
\nonumber b_1' =2\lambda_1(\theta_*)-\frac{1}{K}, c_1' =\frac{1+a_4+(1+a_2^2)(b-1) }{b}\\
\nonumber  c_2'=\frac{1}{\beta_1}\Big ( \beta_3^2 + (\beta_2 \cdot a_1)^2 \cdot (b-1) + (\beta_2 + (b-1)\cdot \beta_1^2) \Big ) , c_3' =K \cdot \beta_1^2\\
\text{ and }  \gamma > \max \left ( \frac{b_1'^2}{c_1'},\frac{\epsilon^2 \delta + \theta_*^2 \cdot \frac{c_2'}{c_1}}{\epsilon^2 \delta    - \theta_*^2 \cdot \frac{c_3'}{b_1} } \right).
\end{gather} 
\end{theorem}

\begin{remark}{We collate the following salient points about the structure of  Theorem  \ref{thm:dadushrelu:noise} : } 

{\bf (a) } Note that for any fixed $\delta$, the $\epsilon$ error guaranteed by the theorem  approaches $0$ as $\sup_{\x} \beta  (\x) \to 0$. Thus we have continuous improvement of the minimum achievable error as the likelihood of the data-poisoning attack decreases. 

{\bf (b)} $\norm{\w_{\rm T} - \w_*}^2  \leq \epsilon^2 \implies \E_{\x} 
\Big [ \Big ( \relu(\w_{\rm T}^\top \x)- \relu(\w_{*}^\top \x) \Big )^2\Big] \leq \epsilon^2 \E \Big [ \norm{\x}^2 \Big ]$ and hence Algorithm \ref{dadushrelu:noise} solves the risk minimization problem for $\theta =0$ to any desired accuracy and in linear time. 

{\bf (c)} Note that the above convergence holds starting from an arbitrary initialization $\w_1$.

{\bf (d)} In subsection \ref{opt:dadush} we shall see how the above theorem gives a {\it worst-case} near-optimal trade-off between $\epsilon$ (the accuracy ) and $\delta$ (the confidence) that can be achieved when training against a $\theta^*$ (a \textit{constant}) additive norm bounded adversary corrupting the true output.

{\bf (e)} \textit{Convergence speed increases with the minibatch size $b$ :}

In the Case (I) above i.e when $\theta_*=0$, one can read off from the proof that upon defining $b_1=2 \lambda_1 ~\& ~c_1=\frac{a_4+a_2^2(b-1)}{b}$, one can find $\delta_0$ so that $c_1> \frac{b_1^2 \delta_0}{(1+\delta_0)^2}$ and upon choosing $\eta= b_1/(c_1(1+\delta_0))$ we obtain 

\[ {\rm T} = 1+\left( \frac{\log \frac{\norm{\w_1 - \w_*}^2}{\epsilon^2 \delta}}{\log \frac{1}{\alpha}}  \right) \] 

\[  \text{ where } \alpha =  1-\frac{4\lambda_1^2\delta_0}{\left ( a_2^2 + \frac{(a_4 -a_2^2)}{b}\right ) \cdot (1+\delta_0)^2}\]

Note that this ${\rm T}$ is a decreasing function of the batchsize $b$ and hence quantifies the intuition that to achieve a pre-specified level of precision, it takes lesser time when using larger batch-sizes. 

A similar conclusion prevails in the $\theta_*>0$ case as well.

{\bf (f)} \textit{The distributional condition is mild :}

Corresponding to both the situations, $\theta_*=0$ and $\theta_*>0$, here we provide simple examples that satisfy the condition of $\lambda_1(\theta_*) >0$.

Example 1: Compact multivariate distribution

Suppose $n=2$ and $\x \sim {\rm Unif} [-1,1] \times [-1,1]$ and suppose $\w_*=(-1,1)$. Hence we can define, 

\begin{align*}
d_1(\theta_*) &\coloneqq \E({\bf 1}_{-x_1+x_2>2 \theta_*}x_1^2)= \E({\bf 1}_{x_1+x_2>2 \theta_*}x_2^2)=\frac{1}{48}(7-8\theta_*+(2\theta_*-1)^4)\\
d_2(\theta_*) &\coloneqq \E({\bf 1}_{-x_1+x_2>2\theta_*}x_1x_2) = \frac{1}{32}- \frac{4\theta*}{24}+\frac{4\theta_*^2-1}{16}- \frac{(2\theta_*-1)^4}{32} \\ &\qquad\qquad\qquad\qquad\qquad\qquad +\frac{4\theta_*(2\theta_*-1)}{24}- \frac{(4\theta_*^2-1)(2\theta_*-1)^2}{16}\\ 
\end{align*}

Then we have $\lambda_1 (\theta_*) \coloneqq  \lambda_{\min}\Bigg(\mathbb{E}_{\x \sim {\cal D}}\Big[ {\bf 1}_{-x_1+x_2> 2\theta_*} \x  \x^\top \Big]\Bigg) =d_1(\theta_*)- \abs{d_2(\theta_*)}$ 

Hence ensuring convergence needs, $d_1(\theta_*)>|d_2(\theta_*)|$ and this is satisfied for examples such as : (a) $\theta_*=0$, $\lambda_1 (0)=\frac{1}{6}-0=\frac{1}{6}$ (b) $\theta_*=1$, $\lambda_1 (1)=\frac{1}{16}-\frac{5}{96}=\frac{1}{96}$.

Example 2: Non-compact univariate distribution

Suppose $n=1, x \sim {\cal N}(0,1)$. Then for any $w_*$ we have, 
$$0<\lambda_1(\theta_*)= \E({\bf 1}_{w_* x >2 \theta_*}x^2) \leq \int_{-\infty}^{\infty}x^2\phi(x) dx= 1$$

where $\phi(x)$ is the standard normal p.d.f. This implies $\lambda_1(\theta_*)$ is finite and positive and thus convergence is ensured.

It is easy to demonstrate further examples in other univariate/multivariate and compact/non-compact distributions as well and see that the convergence conditions are not very strong. 
\end{remark}

\subsection{Experimental demonstration of Algorithm \ref{dadushrelu:noise}}\label{sec:exp_relu}

For experiments we sample the data $\x_{t_i}$ (Algorithm \ref{dadushrelu:noise}) in i.i.d fashion from a standard normal distribution in $n=500$ dimensions. We instantiate a data-poisoning attack consistent with the assumptions in Theorem \ref{thm:dadushrelu:noise} in the following way : at the $t^{th}$ iterate we choose $\xi_{t_i} = \theta_*\ind{i \mod 2 = 0} - \theta_*\ind{i \mod 2 \neq 0}$ and $\alpha_{t_i} $ is $0/1$ w.p $\beta \in [0,1]$ for $i = 1,\ldots,b$. 

\begin{figure} [!ht]  
\centering
\includegraphics[width=10cm,height=4.5cm]{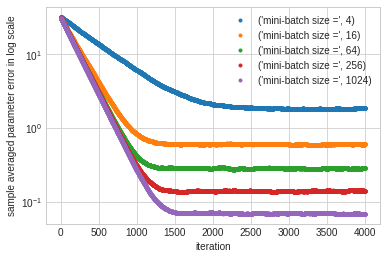}
\caption{Performance of Algorithm \ref{dadushrelu:noise} with changing mini-batch size for $n=500,\beta=0.5$ and $\theta_* =2$}
\label{fig:vsb} 
\end{figure}

\begin{figure} [!ht]  
\centering
\includegraphics[width=10cm,height=4.5cm]{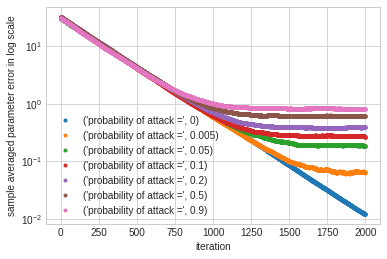}
\caption{Performance of Algorithm \ref{dadushrelu:noise} with changing probability of attack for $n = 500, \theta_* = 2$ and $b=16$}
\label{fig:vsbeta} 
\end{figure}

\begin{figure}[!ht]  
\centering
\includegraphics[width=10cm,height=4.5cm]{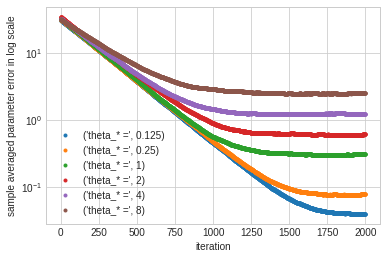}
\caption{Performance of Algorithm \ref{dadushrelu:noise} with changing $\theta_*$ for $n=500,\beta=0.5$ and $b=16$}
\label{fig:vstheta} 
\end{figure}

Then for a chosen value of $\w_*$ and $\eta = 0.01$, we plot how the parameter recovery error $\norm{\w_t - \w_*}$ (averaged over multiple runs of the algorithm) varies with $t$,

\begin{itemize}
    \item for different values of $b$, at fixed $\theta_* = 2$ and $\beta = 0.5$ in Figure \ref{fig:vsb}. Here we can see that larger values of mini-batch help attain lower errors faster. 
    \item for different values of $\beta$, at fixed $\theta_* = 2$ and $b = 16$ in Figure \ref{fig:vsbeta}. Here we can see that there is a graceful degradation of the best achieved error with increasing probability of attack. 
    \item for different values of $\theta_*$, at fixed $\beta = 0.5$ and $b=16$ in Figure \ref{fig:vstheta}. Here we can see that there is a graceful degradation of the best achieved error with increasing magnitude of the attack.  
\end{itemize} 

We note that all the three observations above are consistent with what we would have expected from Theorem \ref{thm:dadushrelu:noise}. 

\begin{figure} [!ht]  
\centering
\includegraphics[width=10cm,height=4.5cm]{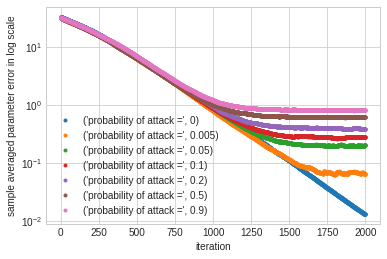}
\caption{Performance of S.G.D. with changing probability of attack for $n = 500, \theta_* = 2$ and $b=16$}
\label{fig:S.G.D.vsbeta} 
\end{figure}

\begin{figure} [!ht]  
\centering
\includegraphics[width=10cm,height=4.5cm]{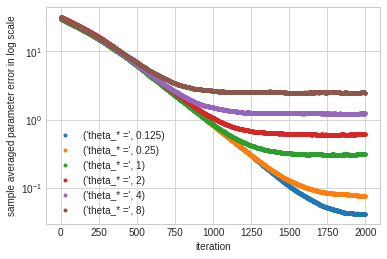}
\caption{Performance of S.G.D. with changing $\theta_*$ for $n=500,\beta=0.5$ and $b=16$}
\label{fig:S.G.D.vstheta} 
\end{figure}

We recall that in  Algorithm \ref{dadushrelu:noise} if we redefined $\g_t$ to, $- \frac{1}{b} \sum_{i=1}^b \ind{\w_{t}^\top \x_{t_i} > 0} (y_{t_i} - \w_t^\top \x_{t_i})\x_{t_i}$ then it would be standard S.G.D. For comparison, we repeat the last two experiments with this  S.G.D. and give the corresponding plots in Figure \ref{fig:S.G.D.vsbeta} and Figure \ref{fig:S.G.D.vstheta}.

We notice the striking similarity between the plots in Figures \ref{fig:vsbeta} \& \ref{fig:S.G.D.vsbeta} and Figures \ref{fig:vstheta} \& \ref{fig:S.G.D.vstheta} respectively. This motivates that our algorithm very closely mimics the behaviour of S.G.D. while similar guarantees as in Theorem \ref{thm:dadushrelu:noise} yet remain elusive for S.G.D..

\subsection{Near-optimality of Theorem~\ref{thm:dadushrelu:noise}}\label{opt:dadush}

We consider the ``worst case" situation of Theorem~\ref{thm:dadushrelu:noise} i.e when $\beta =1$ identically and hence the adversary always acts.  Now consider another value for the filter $\R^r \ni \w_{{\rm adv}} \neq \w^*$ being chosen by this adversary and suppose that $\theta^* = \theta_{\rm adv}$ s.t

\begin{align}\label{thetaopt}
\theta_{\rm adv} \geq \sup_{\x \in \text{supp}(\mathcal{D})} \vert \relu(\w_{\rm adv}^\top \x) - \relu(\w_*^\top \x) \vert
\end{align}

It is easy to imagine cases where the supremum in the RHS above exists like when ${\cal D}$ is compactly supported. Now in this situation we define $\c_{\rm bound } \coloneqq \frac{(2 \lambda_1(\theta_*)- \frac{1}{K})}{\beta_1^2 \cdot K }$ and hence Theorem \ref{thm:dadushrelu:noise} says that the lowest value of the parameter error achievable is, 

\begin{align}\label{epsopt}
\epsilon^2 = \frac{\theta^{\star 2}}{\delta c_{\rm bound}} \implies \epsilon^2 \geq \frac{\theta_{\rm adv}^2}{\c_{\rm bound}}
\end{align}

Hence proving the optimality of this guarantee is equivalent to showing the existence of an attack within this $\theta_{\rm adv}$ bound for which the best accuracy possible nearly saturates the lowerbound in equation \ref{epsopt}.

We note that for the choice of corruption bound $\theta_{\rm adv}$, the adversarial oracle when queried with $\x$ can respond with $\xi_\x + \relu(\w_*^\top \x)$ where $\xi_\x = \relu(\w_{\rm adv}^\top\x) - \relu(\w_*^\top \x)$. 
Hence the data received by the algorithm can be exactly realized with the filter choice $\w_{\rm adv}$.
In that case, the analysis of Theorem \ref{thm:dadushrelu:noise}, Case (I) shows that Algorithm \ref{dadushrelu:noise} will converge in high probability to $\w_{\rm adv}$.
Thus the error incurred is $\epsilon \geq  \norm{\w_{\rm adv} - \w_*}$.

An instantiation of the above attack happening is when $\theta_{\rm adv} = r\norm{\w_{\rm adv} - \w_*}$ for $r = \sup_{\x \in {\rm supp} ({\cal D})} \norm{\x}$. Its easy to imagine cases where ${\cal D}$ is s.t $r$ defined above is finite. Further, this choice of $\theta_{\rm adv}$ is valid since the following holds, as required by equation \ref{thetaopt},

\begin{align*} 
\sup_{\x \in \text{supp}(\mathcal{D})} \vert \relu(\w_{\rm adv}^\top \x) - \relu(\w_*^\top \x ) \vert \leq r\norm{\w_{\rm adv} - \w_*} = \theta_{\rm adv}
\end{align*}

Thus the above setup invoked on training a $\relu$ gate with inputs being sampled from ${\cal D}$ as above while the labels are being additively corrupted by at most $\theta_* (=\theta_{\rm adv}) = r\norm{\w_{\rm adv} - \w_*}$ demonstrates a case where the {\it worst case} accuracy guarantee of $\epsilon^2 \geq \frac{\theta_{\rm adv}^2}{c_{\rm bound}}$ is optimal up to a constant $\frac{r^2}{c_{\rm bound}}$. We note that this argument also implies the worst-case near optimality of guarantees like equation \ref{epsopt} for {\it any} algorithm defending against this attack which also has the property of recovering the parameters correctly when the labels are exactly realizable.

\section{Conclusion} 
In this work we have shown provable training of a $\relu$ gate under mild distributional conditions and pointed out cases where this happens in linear time while assuming only certain mild non-degeneracy conditions on the distribution. Also our results have probed how closely we can recover the original generating weights when the true training labels are subject to an (online) data-poisoning attack.  And in this particular regime, in Section \ref{sec:exp_relu}, we have given careful experimental evidence as to how our provably convergent modification of S.G.D. on a $\relu$ gate (Algorithm \ref{dadushrelu:noise}) seems to have very similar time dynamics as S.G.D. - while for the later such guarantees remain unknown. 

We believe this raises the interesting question as to whether indeed one can rigorously show that the stochastic process induced by Algorithm \ref{dadushrelu:noise}, is a close approximant of true S.G.D. on a $\relu$ gate. We posit that this is a fruitful direction for future investigations and might lead to insights about the dynamics of S.G.D. for nets with a constant number of gates, which has so far mostly remained out of current mathematical reach.  

\section{Acknowledgement}
We are thankful to the Co-Editor in Chief, Action Editor, and referees at the ``Neural Networks" journal for their constructive comments which have significantly helped towards this final form of the paper. Sayar Karmakar's research is partially supported by NSF DMS 2124222. Anirbit Mukherjee would like to thank the inaugural MINDS Data Science Fellowship at J.H.U., Wharton Dean’s Fund for Postdoctoral Research and Weijie Su's NSF CAREER DMS-1847415 for funding this research. 

We would like to thank Daniel Dadush for his critical insights which led to the initial version of the Algorithm \ref{dadushrelu:noise} (which first appeared in \cite{thesis}). Multiple discussions with Amitabh Basu and Anup Rao (during Anirbit's internship at Adobe, San Jose) helped shape the core questions that were pursued in this paper. We would also like to acknowledge the collaboration with Ramchandran Muthukumar during the initial stages of the project. 


\clearpage 
\bibliographystyle{elsarticle-num-names} 
\bibliography{main}

\begin{thebibliography}{41}
\expandafter\ifx\csname natexlab\endcsname\relax\def\natexlab#1{#1}\fi
\providecommand{\url}[1]{\texttt{#1}}
\providecommand{\href}[2]{#2}
\providecommand{\path}[1]{#1}
\providecommand{\DOIprefix}{doi:}
\providecommand{\ArXivprefix}{arXiv:}
\providecommand{\URLprefix}{URL: }
\providecommand{\Pubmedprefix}{pmid:}
\providecommand{\doi}[1]{\href{http://dx.doi.org/#1}{\path{#1}}}
\providecommand{\Pubmed}[1]{\href{pmid:#1}{\path{#1}}}
\providecommand{\bibinfo}[2]{#2}
\ifx\xfnm\relax \def\xfnm[#1]{\unskip,\space#1}\fi
\bibitem[{Silver et~al.(2016)Silver, Huang, Maddison, Guez, Sifre, Van
  Den~Driessche, Schrittwieser, Antonoglou, Panneershelvam, Lanctot
  et~al.}]{silver2016mastering}
\bibinfo{author}{D.~Silver}, \bibinfo{author}{A.~Huang}, \bibinfo{author}{C.~J.
  Maddison}, \bibinfo{author}{A.~Guez}, \bibinfo{author}{L.~Sifre},
  \bibinfo{author}{G.~Van Den~Driessche}, \bibinfo{author}{J.~Schrittwieser},
  \bibinfo{author}{I.~Antonoglou}, \bibinfo{author}{V.~Panneershelvam},
  \bibinfo{author}{M.~Lanctot}, et~al.,
\newblock \bibinfo{title}{Mastering the game of go with deep neural networks
  and tree search},
\newblock \bibinfo{journal}{nature} \bibinfo{volume}{529}
  (\bibinfo{year}{2016}) \bibinfo{pages}{484--489}.
\bibitem[{Silver et~al.(2017)Silver, Schrittwieser, Simonyan, Antonoglou,
  Huang, Guez, Hubert, Baker, Lai, Bolton et~al.}]{silver2017mastering}
\bibinfo{author}{D.~Silver}, \bibinfo{author}{J.~Schrittwieser},
  \bibinfo{author}{K.~Simonyan}, \bibinfo{author}{I.~Antonoglou},
  \bibinfo{author}{A.~Huang}, \bibinfo{author}{A.~Guez},
  \bibinfo{author}{T.~Hubert}, \bibinfo{author}{L.~Baker},
  \bibinfo{author}{M.~Lai}, \bibinfo{author}{A.~Bolton}, et~al.,
\newblock \bibinfo{title}{Mastering the game of go without human knowledge},
\newblock \bibinfo{journal}{Nature} \bibinfo{volume}{550}
  (\bibinfo{year}{2017}) \bibinfo{pages}{354--359}.
\bibitem[{Silver et~al.(2018)Silver, Hubert, Schrittwieser, Antonoglou, Lai,
  Guez, Lanctot, Sifre, Kumaran, Graepel et~al.}]{silver2018general}
\bibinfo{author}{D.~Silver}, \bibinfo{author}{T.~Hubert},
  \bibinfo{author}{J.~Schrittwieser}, \bibinfo{author}{I.~Antonoglou},
  \bibinfo{author}{M.~Lai}, \bibinfo{author}{A.~Guez},
  \bibinfo{author}{M.~Lanctot}, \bibinfo{author}{L.~Sifre},
  \bibinfo{author}{D.~Kumaran}, \bibinfo{author}{T.~Graepel}, et~al.,
\newblock \bibinfo{title}{A general reinforcement learning algorithm that
  masters chess, shogi, and go through self-play},
\newblock \bibinfo{journal}{Science} \bibinfo{volume}{362}
  (\bibinfo{year}{2018}) \bibinfo{pages}{1140--1144}.
\bibitem[{Schrittwieser et~al.(2020)Schrittwieser, Antonoglou, Hubert,
  Simonyan, Sifre, Schmitt, Guez, Lockhart, Hassabis, Graepel
  et~al.}]{schrittwieser2020mastering}
\bibinfo{author}{J.~Schrittwieser}, \bibinfo{author}{I.~Antonoglou},
  \bibinfo{author}{T.~Hubert}, \bibinfo{author}{K.~Simonyan},
  \bibinfo{author}{L.~Sifre}, \bibinfo{author}{S.~Schmitt},
  \bibinfo{author}{A.~Guez}, \bibinfo{author}{E.~Lockhart},
  \bibinfo{author}{D.~Hassabis}, \bibinfo{author}{T.~Graepel}, et~al.,
\newblock \bibinfo{title}{Mastering atari, go, chess and shogi by planning with
  a learned model},
\newblock \bibinfo{journal}{Nature} \bibinfo{volume}{588}
  (\bibinfo{year}{2020}) \bibinfo{pages}{604--609}.
\bibitem[{Neal(1996)}]{neal1996priors}
\bibinfo{author}{R.~M. Neal},
\newblock \bibinfo{title}{Priors for infinite networks},
\newblock in: \bibinfo{booktitle}{Bayesian Learning for Neural Networks},
  \bibinfo{publisher}{Springer}, \bibinfo{year}{1996}, pp.
  \bibinfo{pages}{29--53}.
\bibitem[{Chizat and Bach(2018)}]{chizat2018global}
\bibinfo{author}{L.~Chizat}, \bibinfo{author}{F.~Bach},
\newblock \bibinfo{title}{On the global convergence of gradient descent for
  over-parameterized models using optimal transport},
\newblock in: \bibinfo{booktitle}{Advances in neural information processing
  systems}, \bibinfo{year}{2018}, pp. \bibinfo{pages}{3036--3046}.
\bibitem[{Jacot et~al.(2018)Jacot, Gabriel, and Hongler}]{jacot2018neural}
\bibinfo{author}{A.~Jacot}, \bibinfo{author}{F.~Gabriel},
  \bibinfo{author}{C.~Hongler},
\newblock \bibinfo{title}{Neural tangent kernel: Convergence and generalization
  in neural networks},
\newblock in: \bibinfo{booktitle}{Advances in neural information processing
  systems}, \bibinfo{year}{2018}, pp. \bibinfo{pages}{8571--8580}.
\bibitem[{Lee et~al.(2017)Lee, Bahri, Novak, Schoenholz, Pennington, and
  Sohl-Dickstein}]{lee2017deep}
\bibinfo{author}{J.~Lee}, \bibinfo{author}{Y.~Bahri},
  \bibinfo{author}{R.~Novak}, \bibinfo{author}{S.~S. Schoenholz},
  \bibinfo{author}{J.~Pennington}, \bibinfo{author}{J.~Sohl-Dickstein},
  \bibinfo{title}{Deep neural networks as gaussian processes},
  \bibinfo{year}{2017}. \href{http://arxiv.org/abs/1711.00165}{{\tt
  arXiv:1711.00165}}.
\bibitem[{Wu et~al.(2019)Wu, Du, and Ward}]{wu2019global}
\bibinfo{author}{X.~Wu}, \bibinfo{author}{S.~S. Du}, \bibinfo{author}{R.~Ward},
\newblock \bibinfo{title}{Global convergence of adaptive gradient methods for
  an over-parameterized neural network},
\newblock \bibinfo{journal}{arXiv preprint arXiv:1902.07111}
  (\bibinfo{year}{2019}).
\bibitem[{Du et~al.(2018)Du, Lee, Li, Wang, and Zhai}]{du2018gradient}
\bibinfo{author}{S.~S. Du}, \bibinfo{author}{J.~D. Lee},
  \bibinfo{author}{H.~Li}, \bibinfo{author}{L.~Wang},
  \bibinfo{author}{X.~Zhai}, \bibinfo{title}{Gradient descent finds global
  minima of deep neural networks}, \bibinfo{year}{2018}.
  \href{http://arxiv.org/abs/1811.03804}{{\tt arXiv:1811.03804}}.
\bibitem[{Su and Yang(2019)}]{su2019learning}
\bibinfo{author}{L.~Su}, \bibinfo{author}{P.~Yang},
\newblock \bibinfo{title}{On learning over-parameterized neural networks: A
  functional approximation perspective},
\newblock in: \bibinfo{booktitle}{Advances in Neural Information Processing
  Systems}, \bibinfo{year}{2019}, pp. \bibinfo{pages}{2637--2646}.
\bibitem[{Kawaguchi and Huang(2019)}]{kawaguchi2019gradient}
\bibinfo{author}{K.~Kawaguchi}, \bibinfo{author}{J.~Huang},
\newblock \bibinfo{title}{Gradient descent finds global minima for
  generalizable deep neural networks of practical sizes},
\newblock in: \bibinfo{booktitle}{2019 57th Annual Allerton Conference on
  Communication, Control, and Computing (Allerton)},
  \bibinfo{organization}{IEEE}, \bibinfo{year}{2019}, pp.
  \bibinfo{pages}{92--99}.
\bibitem[{Huang and Yau(2019)}]{huang2019dynamics}
\bibinfo{author}{J.~Huang}, \bibinfo{author}{H.-T. Yau},
\newblock \bibinfo{title}{Dynamics of deep neural networks and neural tangent
  hierarchy},
\newblock \bibinfo{journal}{arXiv preprint arXiv:1909.08156}
  (\bibinfo{year}{2019}).
\bibitem[{Allen-Zhu et~al.(2019{\natexlab{a}})Allen-Zhu, Li, and
  Song}]{allen2019convergenceDNN}
\bibinfo{author}{Z.~Allen-Zhu}, \bibinfo{author}{Y.~Li},
  \bibinfo{author}{Z.~Song},
\newblock \bibinfo{title}{A convergence theory for deep learning via
  over-parameterization},
\newblock in: \bibinfo{booktitle}{International Conference on Machine
  Learning}, \bibinfo{year}{2019}{\natexlab{a}}, pp. \bibinfo{pages}{242--252}.
\bibitem[{Allen-Zhu et~al.(2019{\natexlab{b}})Allen-Zhu, Li, and
  Liang}]{allen2019learning}
\bibinfo{author}{Z.~Allen-Zhu}, \bibinfo{author}{Y.~Li},
  \bibinfo{author}{Y.~Liang},
\newblock \bibinfo{title}{Learning and generalization in overparameterized
  neural networks, going beyond two layers},
\newblock in: \bibinfo{booktitle}{Advances in neural information processing
  systems}, \bibinfo{year}{2019}{\natexlab{b}}, pp.
  \bibinfo{pages}{6155--6166}.
\bibitem[{Allen-Zhu et~al.(2019{\natexlab{c}})Allen-Zhu, Li, and
  Song}]{allen2019convergenceRNN}
\bibinfo{author}{Z.~Allen-Zhu}, \bibinfo{author}{Y.~Li},
  \bibinfo{author}{Z.~Song},
\newblock \bibinfo{title}{On the convergence rate of training recurrent neural
  networks},
\newblock in: \bibinfo{booktitle}{Advances in Neural Information Processing
  Systems}, \bibinfo{year}{2019}{\natexlab{c}}, pp.
  \bibinfo{pages}{6673--6685}.
\bibitem[{Du and Lee(2018)}]{du2018power}
\bibinfo{author}{S.~Du}, \bibinfo{author}{J.~Lee},
\newblock \bibinfo{title}{On the power of over-parametrization in neural
  networks with quadratic activation},
\newblock in: \bibinfo{booktitle}{International Conference on Machine
  Learning}, \bibinfo{year}{2018}, pp. \bibinfo{pages}{1329--1338}.
\bibitem[{Zou et~al.(2018)Zou, Cao, Zhou, and Gu}]{zou2018stochastic}
\bibinfo{author}{D.~Zou}, \bibinfo{author}{Y.~Cao}, \bibinfo{author}{D.~Zhou},
  \bibinfo{author}{Q.~Gu},
\newblock \bibinfo{title}{Stochastic gradient descent optimizes
  over-parameterized deep relu networks},
\newblock \bibinfo{journal}{arXiv preprint arXiv:1811.08888}
  (\bibinfo{year}{2018}).
\bibitem[{Zou and Gu(2019)}]{zou2019improved}
\bibinfo{author}{D.~Zou}, \bibinfo{author}{Q.~Gu},
\newblock \bibinfo{title}{An improved analysis of training over-parameterized
  deep neural networks},
\newblock in: \bibinfo{booktitle}{Advances in Neural Information Processing
  Systems}, \bibinfo{year}{2019}, pp. \bibinfo{pages}{2053--2062}.
\bibitem[{Arora et~al.(2019{\natexlab{a}})Arora, Du, Hu, Li, Salakhutdinov, and
  Wang}]{arora2019exact}
\bibinfo{author}{S.~Arora}, \bibinfo{author}{S.~S. Du},
  \bibinfo{author}{W.~Hu}, \bibinfo{author}{Z.~Li}, \bibinfo{author}{R.~R.
  Salakhutdinov}, \bibinfo{author}{R.~Wang},
\newblock \bibinfo{title}{On exact computation with an infinitely wide neural
  net},
\newblock in: \bibinfo{booktitle}{Advances in Neural Information Processing
  Systems}, \bibinfo{year}{2019}{\natexlab{a}}, pp.
  \bibinfo{pages}{8139--8148}.
\bibitem[{Arora et~al.(2019{\natexlab{b}})Arora, Du, Li, Salakhutdinov, Wang,
  and Yu}]{arora2019harnessing}
\bibinfo{author}{S.~Arora}, \bibinfo{author}{S.~S. Du},
  \bibinfo{author}{Z.~Li}, \bibinfo{author}{R.~Salakhutdinov},
  \bibinfo{author}{R.~Wang}, \bibinfo{author}{D.~Yu},
\newblock \bibinfo{title}{Harnessing the power of infinitely wide deep nets on
  small-data tasks},
\newblock \bibinfo{journal}{arXiv preprint arXiv:1910.01663}
  (\bibinfo{year}{2019}{\natexlab{b}}).
\bibitem[{Li et~al.(2019)Li, Wang, Yu, Du, Hu, Salakhutdinov, and
  Arora}]{li2019enhanced}
\bibinfo{author}{Z.~Li}, \bibinfo{author}{R.~Wang}, \bibinfo{author}{D.~Yu},
  \bibinfo{author}{S.~S. Du}, \bibinfo{author}{W.~Hu},
  \bibinfo{author}{R.~Salakhutdinov}, \bibinfo{author}{S.~Arora},
\newblock \bibinfo{title}{Enhanced convolutional neural tangent kernels},
\newblock \bibinfo{journal}{arXiv preprint arXiv:1911.00809}
  (\bibinfo{year}{2019}).
\bibitem[{Arora et~al.(2019)Arora, Du, Hu, Li, and Wang}]{arora2019fine}
\bibinfo{author}{S.~Arora}, \bibinfo{author}{S.~Du}, \bibinfo{author}{W.~Hu},
  \bibinfo{author}{Z.~Li}, \bibinfo{author}{R.~Wang},
\newblock \bibinfo{title}{Fine-grained analysis of optimization and
  generalization for overparameterized two-layer neural networks},
\newblock in: \bibinfo{booktitle}{International Conference on Machine
  Learning}, \bibinfo{year}{2019}, pp. \bibinfo{pages}{322--332}.
\bibitem[{Allen-Zhu and Li(2019)}]{allen2019can}
\bibinfo{author}{Z.~Allen-Zhu}, \bibinfo{author}{Y.~Li},
\newblock \bibinfo{title}{What can resnet learn efficiently, going beyond
  kernels?},
\newblock in: \bibinfo{booktitle}{Advances in Neural Information Processing
  Systems}, \bibinfo{year}{2019}, pp. \bibinfo{pages}{9015--9025}.
\bibitem[{Wei et~al.(2019)Wei, Lee, Liu, and Ma}]{wei2019regularization}
\bibinfo{author}{C.~Wei}, \bibinfo{author}{J.~D. Lee},
  \bibinfo{author}{Q.~Liu}, \bibinfo{author}{T.~Ma},
\newblock \bibinfo{title}{Regularization matters: Generalization and
  optimization of neural nets vs their induced kernel},
\newblock in: \bibinfo{booktitle}{Advances in Neural Information Processing
  Systems}, \bibinfo{year}{2019}, pp. \bibinfo{pages}{9709--9721}.
\bibitem[{Goel et~al.(2016)Goel, Kanade, Klivans, and
  Thaler}]{goel2016reliably}
\bibinfo{author}{S.~Goel}, \bibinfo{author}{V.~Kanade},
  \bibinfo{author}{A.~Klivans}, \bibinfo{author}{J.~Thaler},
\newblock \bibinfo{title}{Reliably learning the relu in polynomial time},
\newblock \bibinfo{journal}{arXiv preprint arXiv:1611.10258}
  (\bibinfo{year}{2016}).
\bibitem[{Frei et~al.(2020)Frei, Cao, and Gu}]{frei2020agnostic}
\bibinfo{author}{S.~Frei}, \bibinfo{author}{Y.~Cao}, \bibinfo{author}{Q.~Gu},
\newblock \bibinfo{title}{Agnostic learning of a single neuron with gradient
  descent},
\newblock \bibinfo{journal}{arXiv preprint arXiv:2005.14426}
  (\bibinfo{year}{2020}).
\bibitem[{Du et~al.(2017)Du, Lee, and Tian}]{dulitian}
\bibinfo{author}{S.~S. Du}, \bibinfo{author}{J.~D. Lee},
  \bibinfo{author}{Y.~Tian},
\newblock \bibinfo{title}{When is a convolutional filter easy to learn?},
\newblock \bibinfo{journal}{arXiv preprint arXiv:1709.06129}
  (\bibinfo{year}{2017}).
\bibitem[{Rosenblatt(1958)}]{rosenblatt1958perceptron}
\bibinfo{author}{F.~Rosenblatt},
\newblock \bibinfo{title}{The perceptron: a probabilistic model for information
  storage and organization in the brain.},
\newblock \bibinfo{journal}{Psychological review} \bibinfo{volume}{65}
  (\bibinfo{year}{1958}) \bibinfo{pages}{386}.
\bibitem[{Pal and Mitra(1992)}]{Pal1992MultilayerPF}
\bibinfo{author}{S.~K. Pal}, \bibinfo{author}{S.~Mitra},
\newblock \bibinfo{title}{Multilayer perceptron, fuzzy sets, and
  classification},
\newblock \bibinfo{journal}{IEEE transactions on neural networks}
  \bibinfo{volume}{3 5} (\bibinfo{year}{1992}) \bibinfo{pages}{683--97}.
\bibitem[{Freund and Schapire(1999)}]{freund1999large}
\bibinfo{author}{Y.~Freund}, \bibinfo{author}{R.~E. Schapire},
\newblock \bibinfo{title}{Large margin classification using the perceptron
  algorithm},
\newblock \bibinfo{journal}{Machine learning} \bibinfo{volume}{37}
  (\bibinfo{year}{1999}) \bibinfo{pages}{277--296}.
\bibitem[{Kakade et~al.(2011)Kakade, Kanade, Shamir, and
  Kalai}]{kakade2011efficient}
\bibinfo{author}{S.~M. Kakade}, \bibinfo{author}{V.~Kanade},
  \bibinfo{author}{O.~Shamir}, \bibinfo{author}{A.~Kalai},
\newblock \bibinfo{title}{Efficient learning of generalized linear and single
  index models with isotonic regression},
\newblock in: \bibinfo{booktitle}{Advances in Neural Information Processing
  Systems}, \bibinfo{year}{2011}, pp. \bibinfo{pages}{927--935}.
\bibitem[{Klivans and Meka(2017)}]{klivans2017learning}
\bibinfo{author}{A.~Klivans}, \bibinfo{author}{R.~Meka},
\newblock \bibinfo{title}{Learning graphical models using multiplicative
  weights},
\newblock in: \bibinfo{booktitle}{2017 IEEE 58th Annual Symposium on
  Foundations of Computer Science (FOCS)}, \bibinfo{organization}{IEEE},
  \bibinfo{year}{2017}, pp. \bibinfo{pages}{343--354}.
\bibitem[{Goel and Klivans(2017)}]{goel2017learning}
\bibinfo{author}{S.~Goel}, \bibinfo{author}{A.~Klivans},
\newblock \bibinfo{title}{Learning depth-three neural networks in polynomial
  time},
\newblock \bibinfo{journal}{arXiv preprint arXiv:1709.06010}
  (\bibinfo{year}{2017}).
\bibitem[{Goel et~al.(2018)Goel, Klivans, and Meka}]{goel2018learning}
\bibinfo{author}{S.~Goel}, \bibinfo{author}{A.~Klivans},
  \bibinfo{author}{R.~Meka},
\newblock \bibinfo{title}{Learning one convolutional layer with overlapping
  patches},
\newblock \bibinfo{journal}{arXiv preprint arXiv:1802.02547}
  (\bibinfo{year}{2018}).
\bibitem[{Johnson and Zhang(2013)}]{johnson2013accelerating}
\bibinfo{author}{R.~Johnson}, \bibinfo{author}{T.~Zhang},
\newblock \bibinfo{title}{Accelerating stochastic gradient descent using
  predictive variance reduction},
\newblock \bibinfo{journal}{Advances in neural information processing systems}
  \bibinfo{volume}{26} (\bibinfo{year}{2013}) \bibinfo{pages}{315--323}.
\bibitem[{Soltanolkotabi(2017)}]{soltanolkotabi2017learning}
\bibinfo{author}{M.~Soltanolkotabi},
\newblock \bibinfo{title}{Learning relus via gradient descent},
\newblock in: \bibinfo{booktitle}{Advances in neural information processing
  systems}, \bibinfo{year}{2017}, pp. \bibinfo{pages}{2007--2017}.
\bibitem[{Kalan et~al.(2019)Kalan, Soltanolkotabi, and
  Avestimehr}]{kalan2019fitting}
\bibinfo{author}{S.~M.~M. Kalan}, \bibinfo{author}{M.~Soltanolkotabi},
  \bibinfo{author}{A.~S. Avestimehr},
\newblock \bibinfo{title}{Fitting relus via sgd and quantized sgd},
\newblock in: \bibinfo{booktitle}{2019 IEEE International Symposium on
  Information Theory (ISIT)}, \bibinfo{organization}{IEEE},
  \bibinfo{year}{2019}, pp. \bibinfo{pages}{2469--2473}.
\bibitem[{Diakonikolas et~al.(2020{\natexlab{a}})Diakonikolas, Goel, Karmalkar,
  Klivans, and Soltanolkotabi}]{diakonikolas2020approximation}
\bibinfo{author}{I.~Diakonikolas}, \bibinfo{author}{S.~Goel},
  \bibinfo{author}{S.~Karmalkar}, \bibinfo{author}{A.~R. Klivans},
  \bibinfo{author}{M.~Soltanolkotabi},
\newblock \bibinfo{title}{Approximation schemes for relu regression},
\newblock in: \bibinfo{booktitle}{Conference on Learning Theory},
  \bibinfo{year}{2020}{\natexlab{a}}.
\bibitem[{Diakonikolas et~al.(2020{\natexlab{b}})Diakonikolas, Kontonis,
  Tzamos, and Zarifis}]{diakonikolas2020learning}
\bibinfo{author}{I.~Diakonikolas}, \bibinfo{author}{V.~Kontonis},
  \bibinfo{author}{C.~Tzamos}, \bibinfo{author}{N.~Zarifis},
\newblock \bibinfo{title}{Learning halfspaces with massart noise under
  structured distributions},
\newblock \bibinfo{journal}{arXiv preprint arXiv:2002.05632}
  (\bibinfo{year}{2020}{\natexlab{b}}).
\bibitem[{Mukherjee(2021)}]{thesis}
\bibinfo{author}{A.~Mukherjee},
\newblock \bibinfo{title}{A study of the mathematics of deep learning},
\newblock \bibinfo{journal}{CoRR} \bibinfo{volume}{abs/2104.14033}
  (\bibinfo{year}{2021}). \URLprefix \url{https://arxiv.org/abs/2104.14033}.
  \href{http://arxiv.org/abs/2104.14033}{{\tt arXiv:2104.14033}}.

\end{thebibliography}

\newcommand{\proofofref}{}
\newproof{zproofof}{Proof of \proofofref}
\newenvironment{proofof}[1]
 {\renewcommand{\proofofref}{#1}\zproofof}
 {\endzproofof}

\appendix 
\section{Proofs of Section \ref{sec:glmtron}}\label{sec:appendixb}
\subsection{Proof of Lemma \ref{stepdecrease}}\label{app:glmtronstep}

\begin{proofof}{Lemma \ref{stepdecrease}}
We observe that, 
\begin{align}\label{main} 
\nonumber &\norm{\w_t - \w}^2 - \norm{\w_{t+1} - \w}^2    = \norm{\w_t - \w}^2 - \norm{\Big ( \w_{t} + \frac {1}{S} \sum_{i=1}^S \Big (y_i - \sigma (\langle \w_t , \x_i\rangle) \Big )\x_i \Big ) - \w}^2\\
\nonumber &= -\frac{2}{S} \sum_{i=1}^S \Big \langle \Big (y_i - \sigma (\langle \w_t , \x_i\rangle) \Big)\x_i, \w_t - \w \Big  \rangle - \norm{\frac{1}{S}\sum_{i=1}^S \Big ( y_i - \sigma (\langle \w_t , \x_i\rangle ) \Big )\x_i}^2\\
&= \frac{2}{S} \sum_{i=1}^S \Big ( y_i - \sigma (\langle \w_t , \x_i\rangle) \Big ) \Big (\langle \w,\x_i \rangle  - \langle \w_t , \x_i \rangle \Big ) - \norm{\frac{1}{S}\sum_{i=1}^S \Big (y_i - \sigma (\langle \w_t , \x_i\rangle) \Big ) \x_i}^2
\end{align}

Analyzing the first term in the RHS above we get, 

\begin{align*} 
&\frac{2}{S} \sum_{i=1}^S \Big ( y_i - \sigma (\langle \w_t , \x_i\rangle) \Big ) \Big (\langle \w,\x_i \rangle  - \langle \w_t , \x_i \rangle \Big )\\
&= \frac{2}{S} \sum_{i=1}^S \Big (y_i - \sigma (\langle \w, \x_i \rangle ) + \sigma  (\langle \w, \x_i \rangle) - \sigma (\langle \w_t , \x_i\rangle) \Big ) \Big (\langle \w,\x_i \rangle  - \langle \w_t , \x_i \rangle \Big )\\
&= \frac{2}{S} \sum_{i=1}^S \Big \langle \Big (y_i - \sigma (\langle \w, \x_i \rangle ) \Big )\x_i, \w - \w_t \Big \rangle 
+   \frac{2}{S} \sum_{i=1}^S \Big (\sigma (\langle \w, \x_i \rangle) - \sigma (\langle \w_t , \x_i\rangle) \Big ) \Big ( \langle \x_i, \w \rangle - \langle \x_i , \w_t \rangle \Big )\\
&\geq - 2\eta W + \frac{2}{S} \sum_{i=1}^S  \Big ( \sigma (\langle \w, \x_i \rangle)  - \sigma (\langle \w_t , \x_i\rangle) \Big ) \Big ( \langle \x_i, \w \rangle - \langle \x_i , \w_t \rangle \Big)
\end{align*} 

In the first term above we have invoked the definition of $\eta$ and $W$ given in the lemma. Further since we are given that $\sigma$ is non-decreasing and $L-$Lipschitz, we have for the second term in the RHS above, 

\begin{align*} 
&\frac{2}{S} \sum_{i=1}^S  \Big ( \sigma (\langle \w, \x_i \rangle)  - \sigma (\langle \w_t , \x_i\rangle) \Big ) \Big ( \langle \x_i, \w \rangle - \langle \x_i , \w_t \rangle \Big) \\
&\geq \frac{2}{SL} \sum_{i=1}^S  \Big ( \sigma (\langle \w, \x_i \rangle) - \sigma (\langle \w_t , \x_i\rangle) \Big )^2 =: \frac{2}{L} \tilde{L}_S (h_t)
\end{align*}

Thus together we have, 

\begin{eqnarray}\label{main1}
\frac{2}{S} \sum_{i=1}^S \Big (y_i - \sigma (\langle \w_t , \x_i\rangle) \Big ) \Big (\langle \w,\x_i \rangle  - \langle \w_t , \x_i \rangle \Big ) \geq -2\eta W  + \frac{2}{L} \tilde{L}_S (h_t) 
\end{eqnarray}

Now we look at the second term in the RHS of equation \ref{main} and that gives us, 

\begin{align}\label{main2}
\nonumber &\norm{\frac{1}{S}\sum_{i=1}^S \Big (y_i - \sigma (\langle \w_t , \x_i\rangle) \Big )\x_i}^2 = \norm{\frac{1}{S}\sum_{i=1}^S \Big (y_i - \sigma (\langle \w , \x_i\rangle) + \sigma(\langle \w , \x_i\rangle) - \sigma(\langle \w_t , \x_i\rangle ) \Big )\x_i}^2\\
\nonumber &\leq \norm{\frac{1}{S}\sum_{i=1}^S \Big (y_i - \sigma (\langle \w , \x_i\rangle) \Big )\x_i}^2\\
\nonumber &+ 2 \norm{\frac{1}{S}\sum_{i=1}^S \Big (y_i - \sigma(\langle \w , \x_i\rangle) \Big)\x_i} \times \norm{\frac{1}{S}\sum_{i=1}^S \Big (\sigma(\langle \w , \x_i\rangle) - \sigma(\langle \w_t , \x_i\rangle) \Big)\x_i}\\
\nonumber &+ \norm{\frac{1}{S}\sum_{i=1}^S \Big (\sigma(\langle \w , \x_i\rangle) - \sigma(\langle \w_t , \x_i\rangle) \Big )\x_i}^2\\
 &\leq \eta^2 + 2\eta  \norm{\frac{1}{S}\sum_{i=1}^S \Big (\sigma(\langle \w , \x_i\rangle) - \sigma(\langle \w_t , \x_i\rangle) \Big)\x_i} + \norm{\frac{1}{S}\sum_{i=1}^S \Big (\sigma(\langle \w , \x_i\rangle) - \sigma(\langle \w_t , \x_i\rangle) \Big )\x_i}^2
\end{align}

Now by Jensen's inequality we have, 

\begin{align*}
&\norm{\frac{1}{S}\sum_{i=1}^S \Big (\sigma(\langle \w , \x_i\rangle) - \sigma(\langle \w_t , \x_i\rangle) \Big)\x_i}^2
\leq \frac{1}{S}\sum_{i=1}^S \Big (\sigma(\langle \w , \x_i\rangle) - \sigma(\langle \w_t , \x_i\rangle) \Big )^2 = \tilde{L}_S(h_t)
\end{align*}

And we have from the definition of $L$ and $W$,

\[ \norm{\frac{1}{S}\sum_{i=1}^S \Big ( \sigma(\langle \w , \x_i\rangle) - \sigma(\langle \w_t , \x_i\rangle) \Big )\x_i} \leq \frac{L}{S}\sum_{i=1}^S \norm{ \w - \w_t} \leq L \times W  \].

Substituting the above two into the RHS of equation \ref{main2} we have,

\begin{align}\label{main3} 
\norm{\frac{1}{S}\sum_{i=1}^S \Big (y_i - \sigma (\langle \w_t , \x_i\rangle) \Big )\x_i}^2 \leq \eta^2 + 2\eta LW + \tilde{L}_S(h_t)
\end{align} 

Now we substitute equations \ref{main1} and \ref{main3} into equation \ref{main} to get, 

\[  \norm{\w_t - \w}^2 - \norm{\w_{t+1} - \w}^2  \geq \Big ( -2\eta W  + \frac{2}{L} \tilde{L}_S (h_t) \Big ) -(\eta^2 + 2\eta LW + \tilde{L}_S(h_t))    \] 

The above simplifies to the inequality we claimed in the lemma i.e,

\[ \norm{\w_{t+1} - \w}^2  \leq  \norm{\w_t - \w}^2 - \Big ( \frac 2 L -1  \Big )\tilde{L}_S (h_t) + \Big ( \eta^2 + 2\eta W (L +1 )  \Big )  \] 
\qed 
\end{proofof}

\subsection{Proof of Theorem \ref{GLMTronReLU}}\label{app:GLMTronReLU} 

\begin{proofof}{Theorem \ref{GLMTronReLU}}
The equation defining the labels in the data-set i.e $y_i = \sigma (\langle \w_*, \x_i \rangle ) + \xi_i$, with $\vert \xi_i \vert \leq \theta$ along with our assumption that, $\norm{\x_i} \leq 1$ implies that , $\norm{\frac {1}{S} \sum_{i=1}^S \Big ( y_i - \sigma (\langle \w_*, \x_i \rangle) \Big )\x_i} \leq \theta$. Thus we can invoke the above Lemma \ref{stepdecrease} between the $t^{th}$ and the $(t+1)^{th}$ iterate with $\w = \w_*$, $\eta = \theta$ and $W = W_t$ s.t $W_t \geq \norm{\w_t - \w} =  \norm{\w_t - \w_*}$ to get,   

\[  \norm{\w_{t+1} - \w_*}^2 \leq \norm{\w_t - \w_*}^2 - \left [   \Big ( \frac {2}{L} - 1 \Big ) \tilde{L}_S(h_t) - (\theta^2 + 2\theta \cdot W_t \cdot  (L + 1) ) \right ] \]

Thus, if $\tilde{L}_S(h_t) \geq \frac{L}{2-L} \Big ( \epsilon +  (\theta^2 + 2\theta \cdot  W_t \cdot (L + 1) ) \Big )$  then, $\norm{\w_{t+1} - \w_*}^2 \leq \norm{\w_t - \w_*}^2 - \epsilon$.  
Thus if the above lowerbound on $\tilde{L}_s(h_t)$ holds in the $t^{th}$ step then at the start of the $(t+1)^{th}$ step we still satisfy, $\norm{\w_{t+1}- \w} < \norm{\w_{t}- \w}$. Since the iterations start with $\w_1=0$, in the first step we can choose $W_1 = \norm{\w_*}$. Now we proceed via induction : from what was argued earlier it follows that if till step $t$ we can keep choosing $W_t = \norm{\w_*}$, then till step $t$ we have reduced the distance to $\w_*$ by ${\cal O}(t \cdot \epsilon)$ and either $\tilde{L}_S(h_t) < \frac{L}{2-L} \Big ( \epsilon +  (\theta^2 + 2\theta \cdot  \norm{\w_*} \cdot (L + 1) ) \Big )$ or in the next step we would have $\norm{\w_{t+1} - \w_*}^2 \leq \norm{\w_t - \w_*}^2 - \epsilon$ and hence the distance to $\w_*$ would decrease further by $\epsilon$. 


But the distance to $\w_*$ is lowerbounded by $0$ and hence in at most $\frac{\norm{\w_*}}{\epsilon}$ steps of the above kind we would have to have attained, 

\[ \tilde{L}_S(h_T) = \frac {1}{S} \sum_{i=1}^S \Big ( \sigma (\langle \w_T, \x_i\rangle ) - \sigma (\langle  \w_* , \x_i\rangle) \Big)^2 < \frac{L}{2-L} \Big ( \epsilon +  (\theta^2 + 2\theta \norm{\w_*} (L + 1) ) \Big ) \]

And that proves the theorem we wanted.   
\qed 
\end{proofof}

\subsection{Proof of Theorem \ref{GLMTronReLUNoise}}\label{app:GLMTronReLUNoise}

\begin{proofof}{Theorem \ref{GLMTronReLUNoise}}
Let the true empirical risk at the $T^{th}-$iterate be defined as, 

\[ L_S(h_T) = \frac {1}{S} \sum_{i=1}^S \Big (\sigma(\langle \w_T, \x_i\rangle ) - \sigma (\langle  \w_* , \x_i\rangle) - \xi_i \Big )^2 \] 

Then it follows that,  

\begin{align*}
    &\tilde{L}_S(h_T) - L_S(h_T) = \frac {1}{S} \sum_{i=1}^S \Big ( \sigma (\langle \w_T, \x_i\rangle ) - \sigma (\langle  \w_* , \x_i\rangle) \Big )^2 - \frac {1}{S} \sum_{i=1}^S \Big ( \sigma (\langle \w_T, \x_i\rangle ) - \sigma(\langle  \w_* , \x_i\rangle) - \xi_i \Big)^2\\
    =& \frac {1}{S} \sum_{i=1}^S \xi_i \Big ( -\xi_i + 2 \sigma (\langle \w_T, \x_i\rangle ) - 2 \sigma (\langle  \w_* , \x_i\rangle) \Big )
    = -\frac {1}{S} \sum_{i=1}^S \xi_i^2 +  \frac {2}{S} \sum_{i=1}^S \xi_i \Big ( \sigma (\langle \w_T, \x_i\rangle ) - \sigma (\langle  \w_* , \x_i\rangle) \Big )
\end{align*}

By the assumption of $\xi_i$ being an unbiased noise the second term vanishes when we compute,\\ $\mathbb{E}_{\{ (\x_i , \xi_i) \mid {i=1,\ldots S} \}} \Big [ \tilde{L}_S(h_T) - L_S(h_T) \Big ]$ Thus we are led to,
\[ \mathbb{E}_{\{ (\x_i , \xi_i) \mid {i=1,\ldots S}  \}} \Big [ \tilde{L}_S(h_T) - L_S(h_T) \Big ] = -\frac {1}{m}  \mathbb{E}_{\{ \xi_i\}_{i=1,\ldots S}} \Big [ \sum_{i=1}^m \xi_i^2 \Big ] = -\frac{1}{m} \sum_{i=1}^m \mathbb{E}_{\{ \xi_i\}} \Big [ \xi_i^2 \Big ] = - \E_{\xi} [\xi^2] \] 

For $T = \frac{\norm{\w_*}}{\epsilon}$, we invoke the upperbound on $\tilde{L}_S(h_T)$ from Theorem \ref{GLMTronReLU} and we can combine it with the above to say, 

\[ \mathbb{E}_{\{ (\x_i , \xi_i) \mid {i=1,\ldots S} \}} \Big [ L_S(h_T) \Big ] \leq \E_{\xi} [\xi^2] + \frac{L}{2-L} \Big ( \epsilon +  (\theta^2 + 2\theta \norm{\w_*} (L + 1) ) \Big ) \] 

And this proves the theorem we wanted.  
\qed
\end{proofof}

\section{Proofs of Section \ref{sec:almostSGDReLU}}\label{sec:appendixa}

\subsection{Proof of Theorem \ref{thm:dadushrelu:noise}} 




\begin{proofof}{Theorem \ref{thm:dadushrelu:noise}}

Here we analyze the dynamics of the Algorithm \ref{dadushrelu:noise}. 
\begin{align}
\nonumber &\norm{\w_{t+1}-\w_*}^2 
= \norm{\w_t -\eta\g_t - \w_*}^2 = \norm{\w_t-\w_*}^2 + \eta^2\norm{\g_t}^2 - 2\eta\langle\w_t-\w_*,\g_t\rangle  
\end{align}
Let the training data sampled till the iterate $t$ be $S_t \coloneqq \bigcup_{i=1}^t s_i$. We overload the notation to also denote by $S_t$, the sigma-algebra generated by the samples seen {\it and the $\alpha$s} till the $t$-th iteration. Conditioned on $S_{t-1}$ , $\w_t$ is determined and $g_t$ is random and dependent on the choice of $\s_t$ and $\{\alpha_{t_i},\xi_{t_i} \mid i =1,\ldots,b\}$. We shall denote the collection of random variables $\{\alpha_{t_i} \mid i =1,\ldots,b\}$ as $\alpha_t$. Then taking conditional expectations w.r.t. $S_{t-1}$ of both sides of the above equation we have,  

\begin{align}\label{DBothTerms:b}
 \nonumber & \E_{s_t,\alpha_t} \Bigg [ \norm{\w_{t+1} - \w_*}^2 \bigg| S_{t-1} \Bigg ] \\
 \nonumber &= \E_{s_t,\alpha_t} \Bigg [ \norm{\w_t - \w_*}^2 \bigg| S_{t-1} \Bigg ]
+   \underbrace{2\frac{\eta}{b} \cdot  \sum_{i=1}^b \E_{\x_{t_i},\alpha_{t_i}} \Bigg [ \Big \langle \w_t - \w_* ,  {\bf 1}_{y_{t_i} >\theta_*} \Big (y_{t_i} - \w_t^\top\x_{t_i} \Big ) \x_{t_i} \Big \rangle \bigg| S_{t-1} \Bigg ]}_{\text{Term }1}\\
&+ \underbrace{\eta^2 \E_{\x_{t_i},\alpha_{t_i}} \Big [ \norm{\g_t}^2 \bigg| S_{t-1} \Bigg ]}_{\text{Term } 2}    
\end{align}




Now we simplify the last two terms of the RHS above, starting from the rightmost,

\begin{align}\label{term2:first} &\text{Term } 2 = 
\eta^2 \cdot \mathbb{E} \Bigg [\norm{\g_t}^2 \mid S_{t-1}\Bigg ]\\
\nonumber &= \frac{\eta^2}{b^2} \sum_{i,j=1}^b  \mathbb{E}\Bigg[{\bf 1}_{y_{t_i} >\theta_*} {\bf 1}_{y_{t_j} >\theta_*}  \cdot (y_{t_i} - \w_t^\top \x_{t_i}) \cdot (y_{t_j} - \w_t^\top \x_{t_j}) \cdot \ip{\x_{t_i}}{\x_{t_j}}  \bigg| S_{t-1}\Bigg] \\
\nonumber &= \frac{\eta^2}{b^2} \sum_{i,j=1}^b  \mathbb{E}\Bigg[{\bf 1}_{y_{t_i} >\theta_*} {\bf 1}_{y_{t_j} >\theta_*} \ip{\x_{t_i}}{\x_{t_j}} \cdot  \bigg [\alpha_{t_i}\alpha_{t_j} \xi_{t_i}\xi_{t_j}\\
\nonumber &+  \big( \relu(\w_*^{\top}\x_{t_i} ) - \w_t^\top \x_{t_i} \big) \big( \relu(\w_*^{\top}\x_{t_j} ) - \w_t^\top \x_{t_j} \big)  \\
\nonumber &+ \alpha_{t_i}\xi_{t_i} \big(\relu(\w_*^{\top}\x_{t_j} ) - \w_t^\top \x_{t_j}\big) + \alpha_{t_j}\xi_{t_j} \big(\relu(\w_*^{\top}\x_{t_i} ) - \w_t^\top \x_{t_i}\big)\bigg ]   \bigg| S_{t-1} \Bigg ]\\
\nonumber &\leq \frac{\eta^2}{b^2} \sum_{i,j=1}^b \Bigg ( \mathbb{E}\Bigg[{\bf 1}_{y_{t_i} >\theta_*} {\bf 1}_{y_{t_j} >\theta_*} \abs{\ip{\x_{t_i}}{\x_{t_j}}} \\
\nonumber &\times   \bigg [ \alpha_{t_i}\alpha_{t_j} \theta_*^2 +   \abs{\relu(\w_*^{\top}\x_{t_i} ) - \w_t^\top \x_{t_i} } \cdot  \abs{\relu(\w_*^{\top}\x_{t_j} ) - \w_t^\top \x_{t_j}}\\
\nonumber &+ \theta_* \left (  \alpha_{t_i} \left \lvert \relu(\w_*^{\top}\x_{t_j} ) - \w_t^\top \x_{t_j} \right \rvert  +  \alpha_{t_j} \left \lvert \relu(\w_*^{\top}\x_{t_i} ) - \w_t^\top \x_{t_i} \right \rvert \right ) \bigg ]   \bigg| S_{t-1} \Bigg ] \Bigg)
\end{align}

As events we have for,  $k=i,j, {\bf 1}_{y_{t_k} >\theta_*} \subset {\bf 1}_{\relu(\w_*^\top \x_{t_k}) > 0} = {\bf 1}_{\w_*^\top \x_{t_k} > 0}.$ Hence we can simplify as follows,
\begin{align} \nonumber & \text{Term } 2\\
\nonumber &\leq \frac{\eta^2}{b^2} \sum_{i,j=1}^b \Bigg \{ \mathbb{E}\Bigg[{\bf 1}_{\w_*^\top \x_{t_i} > 0} {\bf 1}_{\w_*^\top \x_{t_j} > 0} \abs{\ip{\x_{t_i}}{\x_{t_j}}}\\
\nonumber &\cdot  \bigg [ \alpha_{t_i}\alpha_{t_j} \theta_*^2 +   \abs{\relu(\w_*^{\top}\x_{t_i} ) - \w_t^\top \x_{t_i} } \cdot  \abs{\relu(\w_*^{\top}\x_{t_j} ) - \w_t^\top \x_{t_j}}\\
\nonumber &+ \theta_* \left (  \alpha_{t_i} \left \lvert \relu(\w_*^{\top}\x_{t_j} ) - \w_t^\top \x_{t_j} \right \rvert  +  \alpha_{t_j} \left \lvert \relu(\w_*^{\top}\x_{t_i} ) - \w_t^\top \x_{t_i} \right \rvert \right ) \bigg ]   \bigg| S_{t-1} \Bigg ] \Bigg \}\\
\nonumber &\leq  \frac{\eta^2}{b^2} \sum_{i,j=1}^b \Bigg \{ \theta_*^2 \cdot \mathbb{E}\Bigg[{\bf 1}_{\w_*^\top \x_{t_i} > 0 } {\bf 1}_{\w_*^\top \x_{t_j} > 0 } \abs{\ip{\x_{t_i}}{\x_{t_j}}} \cdot  \bigg [(\beta(\x_{t_i}) {\bf 1}_{i=j} + \beta(\x_{t_i})\beta(\x_{t_j}){\bf 1}_{i\neq j}) \bigg ]  \bigg| S_{t-1}  \Bigg ]\\
\nonumber &+ {\bf 1}_{i \neq j} \cdot \mathbb{E}\Bigg[{\bf 1}_{\w_*^\top \x_{t_i} > 0} \norm{\x_{t_i}} \cdot  \abs{\w_*^{\top}\x_{t_i}  - \w_t^\top \x_{t_i} } \bigg| S_{t-1} \Bigg ] \times \mathbb{E}\Bigg[{\bf 1}_{\w_*^\top \x_{t_j} > 0} \norm{\x_{t_j}} \cdot  \abs{\w_*^{\top}\x_{t_j}  - \w_t^\top \x_{t_j} } \bigg| S_{t-1} \Bigg ]\\
\nonumber &+ {\bf 1}_{i = j} \cdot \mathbb{E}\Bigg[{\bf 1}_{\w_*^\top \x_{t_i} > 0} \norm{\x_{t_i}}^2 \cdot  \abs{\w_*^{\top}\x_{t_i}  - \w_t^\top \x_{t_i}}^2 \bigg| S_{t-1} \Bigg ] \\
\nonumber &+ \theta_* \cdot {\bf 1}_{i \neq j} \cdot \bigg (  \mathbb{E}\Bigg[{\bf 1}_{\w_*^\top \x_{t_i} > 0} \cdot \beta(\x_{t_i} ) \cdot \norm{\x_{t_i}} \abs{\w_*^{\top}\x_{t_i}  - \w_t^\top \x_{t_i} } \bigg| S_{t-1} \Bigg ] \cdot \mathbb{E}\Bigg[{\bf 1}_{\w_*^\top \x_{t_j} > 0} \norm{\x_{t_j}} \bigg| S_{t-1} \Bigg ] +  (i \leftrightarrow j) \bigg )\\
 &+  2\theta_* \cdot {\bf 1}_{i = j} \cdot \bigg (  \mathbb{E}\Bigg[{\bf 1}_{\w_*^\top \x_{t_i} > 0} \cdot \beta(\x_{t_i} ) \cdot \norm{\x_{t_i}}^2 \abs{\w_*^{\top}\x_{t_i}  - \w_t^\top \x_{t_i} } \bigg| S_{t-1} \Bigg ]   \bigg ) \Bigg \}
\end{align}


In the last inequality above we have used the facts that (a) for $i \neq j$, functions of $\x_{t_i}$ are uncorrelated with functions of $\x_{t_j}$ and (b) that the random variables $\alpha_{t_i}$ and $\alpha_{t_j}$ are independent of each other and of the mini-batch choice $s_t$ and hence they can be replaced by their respective expectations $\beta(\x_{t_i})$ and $\beta(\x_{t_j})$. And for the first term we need to note the $i=j$ case that, $\E [ \alpha_{t_i}^2 ] = \beta(\x_{t_i})$.

Now we can simplify the first term of the RHS of equation \ref{term2:first} as, 

\begin{align*}
&\theta_*^2 \cdot \mathbb{E}\Bigg[{\bf 1}_{y_{t_i} >\theta_*} {\bf 1}_{y_{t_j} >\theta_*} \abs{\ip{\x_{t_i}}{\x_{t_j}}} \cdot  \bigg [(\beta(\x_{t_i}) {\bf 1}_{i=j} + \beta(\x_{t_i})\beta(\x_{t_j}){\bf 1}_{i\neq j}) \bigg ]  \bigg| S_{t-1}  \Bigg ]\\
\leq & \theta_*^2 \cdot  \E_{\x_{t_i}} \Bigg[\beta (\x_{t_i}) \norm{\x_{t_i}}^2 {\bf 1}_{y_{t_i} >\theta_*}  \bigg| S_{t-1} \Bigg ]{\bf 1}_{i=j}\\
&+   \theta_*^2 \cdot  \E_{\x_{t_i}} \Bigg[\beta (\x_{t_i}) \norm{\x_{t_i}} {\bf 1}_{y_{t_i} >\theta_*}  \bigg| S_{t-1}  \Bigg ] \cdot \E_{\x_{t_j}} \Bigg[\beta (\x_{t_j}) \norm{\x_{t_j}} {\bf 1}_{y_{t_j} >\theta_*}   \bigg| S_{t-1} \Bigg ]{\bf 1}_{i\neq j}\\
\leq & \theta_*^2 \cdot  \E_{\x_{t_i}} \Bigg[\beta (\x_{t_i}) \norm{\x_{t_i}}^2 {\bf 1}_{\w_*^\top \x_{t_i} > 0}  \bigg| S_{t-1} \Bigg ]{\bf 1}_{i=j}\\
&+   \theta_*^2 \cdot  \E_{\x_{t_i}} \Bigg[\beta (\x_{t_i}) \norm{\x_{t_i}} {\bf 1}_{\w_*^\top \x_{t_i} > 0}  \bigg| S_{t-1}  \Bigg ] \cdot \E_{\x_{t_j}} \Bigg[\beta (\x_{t_j}) \norm{\x_{t_j}} {\bf 1}_{\w_*^\top \x_{t_j} > 0}   \bigg| S_{t-1} \Bigg ]{\bf 1}_{i\neq j}
\end{align*}

{Since }$\x_{t_i} ~\& ~\x_{t_j}$ are identically distributed, we can invoke the constants,  
$\beta_{1} ~\& ~\beta_2$ and  under taking total expectations the above is bounded by $ \theta_*^2 (\beta_{2}{\bf 1}_{i=j} + \beta_{1}^2{\bf 1}_{i\neq j})$. Using this we have from taking total expectations on both sides of equation \ref{term2:first}, 

\begin{align}\label{term2:second}
&\E \left [ \text{Term } 2 \right ] \leq \frac{\eta^2}{b^2} \cdot \theta_*^2 (b \cdot \beta_{2} + (b^2 - b) \cdot \beta_{1}^2)\\
\nonumber &+ \frac{\eta^2}{b^2} \sum_{i=1}^b \Bigg\{ \E \left [ \mathbb{E}\Bigg[{\bf 1}_{\w_*^\top \x_{t_i} > 0} \norm{\x_{t_i}}^2 \cdot  \abs{\w_*^{\top}\x_{t_i}  - \w_t^\top \x_{t_i}}^2 \bigg| S_{t-1} \Bigg ] \right ]\\
\nonumber &+ 2\theta_*  \cdot \bigg ( \E \left [  \mathbb{E}\Bigg[{\bf 1}_{\w_*^\top \x_{t_i} > 0} \cdot \beta(\x_{t_i} )  \cdot \norm{\x_{t_i}}^2 \abs{\w_*^{\top}\x_{t_i}  - \w_t^\top \x_{t_i} } \bigg| S_{t-1} \Bigg ] \right ]  \bigg ) \Bigg \}\\
\nonumber &+ \frac{\eta^2}{b^2} \sum_{i,j=1, i \neq j}^b \Bigg\{  \E \Bigg[  \mathbb{E}\Bigg[{\bf 1}_{\w_*^\top \x_{t_i} > 0} \norm{\x_{t_i}} \cdot  \abs{\w_*^{\top}\x_{t_i}  - \w_t^\top \x_{t_i} } \bigg| S_{t-1} \Bigg ]\\
\nonumber &\times  \mathbb{E}\Bigg[{\bf 1}_{\w_*^\top \x_{t_j} > 0} \norm{\x_{t_j}} \cdot  \abs{\w_*^{\top}\x_{t_j}  - \w_t^\top \x_{t_j} } \bigg| S_{t-1} \Bigg ] \Bigg]\\
\nonumber &+ \theta_* \cdot \bigg (  \E \left [ \mathbb{E}\Bigg[{\bf 1}_{\w_*^\top \x_{t_i} > 0} \cdot \beta(\x_{t_i} )  \cdot \norm{\x_{t_i}} \abs{\w_*^{\top}\x_{t_i}  - \w_t^\top \x_{t_i} } \bigg| S_{t-1} \Bigg ] \right ] \cdot \mathbb{E}\Bigg[{\bf 1}_{\w_*^\top \x_{t_j} > 0} \norm{\x_{t_j}}  \Bigg ] +  (i \leftrightarrow j) \bigg ) \Bigg\}
\end{align} 

In the last term of the RHS above we have used the fact that conditioned on $S_{t-1}$ a function of $(\w_t,\x_{t_i})$ is uncorrelated with a function of $\x_{t_j}$ for $i \neq j$. Now we further invoke that for $k = i,j$, conditioned on $S_{t-1}$, $\w_t$ is uncorrelated with any function of $\x_{t_k}$ to simplify the above as,

\begin{align}\label{term2:third}
\nonumber &\E \left [ \text{Term } 2 \right ] \leq \frac{\eta^2}{b^2} \cdot \theta_*^2 (b \cdot \beta_{2} + (b^2 - b) \cdot \beta_{1}^2)\\
\nonumber &+ \frac{\eta^2}{b^2} \sum_{i=1}^b \Bigg\{ \E \left [  \norm{\w_*  - \w_t }^2 \right ] \cdot \mathbb{E} \Bigg[ {\bf 1}_{\w_*^\top \x_{t_i} > 0} \norm{\x_{t_i}}^4  \Bigg ] \\
\nonumber &+ 2\theta_*  \cdot \bigg ( \E \left [ \norm{\w_*  - \w_t } \right ] \cdot  \mathbb{E}\Bigg[ {\bf 1}_{\w_*^\top \x_{t_i} > 0} \cdot \beta(\x_{t_i} )  \cdot \norm{\x_{t_i}}^3   \Bigg ]  \bigg ) \Bigg \}\\
\nonumber &+ \frac{\eta^2}{b^2} \sum_{i,j=1, i \neq j}^b \Bigg\{  \E \left [ \norm{\w_*  - \w_t }^2 \right ] \cdot  \mathbb{E}\Bigg[{\bf 1}_{\w_*^\top \x_{t_i} > 0} \norm{\x_{t_i}}^2   \Bigg ]  \times  \mathbb{E}\Bigg[{\bf 1}_{\w_*^\top \x_{t_j} > 0} \norm{\x_{t_j}}^2   \Bigg ] \\
\nonumber &+ \theta_* \cdot \bigg (  \E \left [ \norm{\w_*  - \w_t } \right ] \cdot   \mathbb{E}\Bigg[{\bf 1}_{\w_*^\top \x_{t_i} > 0} \cdot \beta(\x_{t_i} )  \cdot \norm{\x_{t_i}}^2   \Bigg ] \cdot \mathbb{E}\Bigg[{\bf 1}_{\w_*^\top \x_{t_j} > 0} \norm{\x_{t_j}}  \Bigg ] +  (i \leftrightarrow j) \bigg ) \Bigg \}\\
\nonumber &\leq \frac{\eta^2}{b} \cdot \Bigg \{  a_4 \cdot X_t + 2 \theta_* \cdot  \E \left [ \beta_3 \cdot \norm{\w_*  - \w_t } \right ] \Bigg \}\\
&+ \frac{\eta^2}{b^2} \cdot (b^2 - b) \cdot \Bigg \{ a_2^2 \cdot X_t + 2\theta_* \cdot \E \left [ \beta_2 a_1 \cdot \norm{\w_*  - \w_t } \right ] \Bigg  \} + \frac{\eta^2}{b^2} \cdot \theta_*^2 (b \cdot \beta_{2} + (b^2 - b) \cdot \beta_{1}^2)
\end{align}

In the last line above we have recalled that $\x_{t_i}$ and $\x_{t_j}$ are identically distributed and the definitions of $a_1,a_2,a_4,\beta_2 ~\& \beta_3$ and have defined $X_t \coloneqq \E \left [  \norm{\w_*  - \w_t }^2 \right ]$. In the second and the fourth terms of the RHS above we invoke the inequalities,

\[ 2 \theta_* \cdot  \E \left [ \beta_3 \cdot \norm{\w_*  - \w_t } \right ] \leq (\theta_* \cdot \beta_3)^2 + X_t\] 
\[ 2 \theta_* \cdot \E \left [ \beta_2 a_1 \cdot \norm{\w_*  - \w_t } \right ] \leq (\theta_* \cdot \beta_2 \cdot a_1)^2 + X_t\] 

Thus we have, 

\begin{align}\label{term2:fourth}
\nonumber \E \left [ \text{Term } 2 \right ] \leq &\left (  \frac{a_4 +1}{b}  + \frac{(a_2^2 + 1)(b^2 - b)}{b^2} \right ) \cdot \eta^2 \cdot X_t \\
&+ \left (  \frac{(\theta_* \cdot \beta_3)^2}{b} + \frac{ (\theta_* \cdot \beta_2 \cdot a_1)^2(b^2 - b)}{b^2} + \frac{ \theta_*^2 (b \cdot \beta_{2} + (b^2 - b) \cdot \beta_{1}^2)}{b^2} \right ) \cdot \eta^2
\end{align}




\begin{align}\label{term1:first}
\nonumber &\text{Term } 1 = 2\frac{\eta}{b} \cdot  \sum_{i=1}^b \E_{\x_{t_i},\alpha_{t_i}} \Bigg [ \Big \langle \w_t - \w_* ,  {\bf 1}_{y_{t_i} >\theta_*} \Big (y_{t_i} - \w_t^\top\x_{t_i} \Big ) \x_{t_i} \Big \rangle \bigg| S_{t-1} \Bigg ]\\
\nonumber &= 2\frac{\eta}{b} \cdot  \sum_{i=1}^b \E \Bigg [ {\bf 1}_{y_{t_i} >\theta_*} \left (\alpha_{t_i} \xi_{t_i} + \relu(\w_*^{\top}\x_{t_i})  - \w_t^\top\x_{t_i}\right ) \times (\w_t-\w_*)^\top  \x_{t_i}  \bigg| S_{t-1} \Bigg ]\\
\nonumber &\text{Since $\abs{\xi_{t_i}} \leq \theta_*$ it follows that $y_{t_i} > \theta_* \implies \w_*^\top\x_{t_i} > 0$. Hence,}\\
\nonumber &= 2\frac{\eta}{b} \cdot  \sum_{i=1}^b \E \Bigg [ {\bf 1}_{y_{t_i} >\theta_*} \left (\alpha_{t_i} \xi_{t_i} + (\w_* - \w_t)^{\top}\x_{t_i} \right ) \times (\w_t-\w_*)^\top  \x_{t_i}  \bigg| S_{t-1} \Bigg ]\\
\nonumber &= -2\frac{\eta}{b} \cdot  \sum_{i=1}^b \E \Bigg [ {\bf 1}_{y_{t_i} >\theta_*}  (\w_* - \w_t)^{\top}\cdot \x_{t_i} \x_{t_i}^\top \cdot (\w_*-\w_t)    \bigg| S_{t-1} \Bigg ]\\
\nonumber &+ 2\frac{\eta}{b} \cdot  \sum_{i=1}^b \E \Bigg [ {\bf 1}_{y_{t_i} >\theta_*} \cdot \alpha_{t_i} \xi_{t_i} \cdot (\w_t-\w_*)^\top  \x_{t_i}  \bigg| S_{t-1} \Bigg ] \\
\nonumber &\leq -2\frac{\eta}{b} \cdot  \sum_{i=1}^b \lambda_{\min} \Big ( \mathbb{E}\Big[ {\bf 1}_{y_{t_i} > \theta_*} \x_{t_i}  \x_{t_i}^\top \bigg| S_{t-1}  \Big] \Big ) \norm{\w_t - \w_*}^2\\
\nonumber &+ 2\frac{\eta}{b}\cdot \theta_* \cdot  \sum_{i=1}^b   \mathbb{E}\Bigg[ \beta(\x_{t_i}) \cdot {\bf 1}_{y_{t_i} > \theta_*} \cdot  \norm{\x_{t_i}} \bigg| S_{t-1}\Bigg] \cdot  \norm{\w_t-\w_*}\\
\nonumber \implies &\E \left [ \text{Term } 1 \right ] \leq - 2\eta \lambda_1(\theta_*) \cdot X_t + 2 \eta \theta_* \E \left [ \beta_1 \cdot \norm{\w_t-\w_*} \right ]\\
&\leq - 2\eta \lambda_1(\theta_*) \cdot X_t + \eta \Big (K(\theta_* \cdot \beta_1)^2 + \frac{1}{K}X_t \Big ) {\bf 1}_{\theta_*>0}
\end{align}

In the last line above we used the following argument to write the upperbound in terms of $\lambda_1(\theta_*)$ as given in Definition \ref{def:dadush}. We observe that for any $i$, $\mathbb{E}\Bigg[ {\bf 1}_{y_{t_i} > \theta_*} \cdot  \norm{\x_{t_i}} \bigg| S_{t-1}\Bigg] \leq \mathbb{E}\Bigg[ {\bf 1}_{\w_*^\top \x_{t_i} > 0} \cdot  \norm{\x_{t_i}} \bigg| S_{t-1}\Bigg]$. Also note that $y_{t_i} < \theta_* \implies \w_*^\top \x_{t_i} < 2 \theta_*$. Hence for any test vector $\v$ we have,\\
$\v ^\top \left ( \mathbb{E}\Big[ \left ( {\bf 1}_{y_{t_i} > \theta_*} - {\bf 1}_{\w_*^\top \x_{t_i} > 2\theta_*} \right ) \x_{t_i}  \x_{t_i}^\top \bigg| S_{t-1}  \Big]   \right ) \v \geq 0$ and that in turn implies, 

\[ \lambda_{\min} \Big ( \mathbb{E}\Big[ {\bf 1}_{y_{t_i} > \theta_*} \x_{t_i}  \x_{t_i}^\top \bigg| S_{t-1}  \Big] \Big ) \geq \lambda_{\min} \Big ( \mathbb{E}\Big[ {\bf 1}_{\w_*^\top \x_{t_i} > 2\theta_*} \x_{t_i}  \x_{t_i}^\top \bigg| S_{t-1}  \Big] \Big )=\lambda_{\min} \Big ( \mathbb{E}\Big[ {\bf 1}_{\w_*^\top \x_{t_i} > 2\theta_*} \x_{t_i}  \x_{t_i}^\top  \Big] \Big ) \]


\paragraph{Case 1 : $\theta_* =0$}
Taking total expectations on both sides of equation \ref{DBothTerms:b} and setting $\theta_* =0$ in the RHS of equations \ref{term2:third} and \ref{term1:first} we have, 

\begin{align}
    X_{t+1} \leq  \Big (1 - 2 \eta \lambda_1  + \frac{\eta^2}{b} \cdot (a_4 + a_2^2(b-1) ) \Big )X_t
\end{align}

The above recursion is of the same form as analyzed in  Lemma \ref{recurse} with $b_1=2\lambda_1, c_1=\frac{a_4 + a_2^2(b-1)}{b}$ one can see that $c_1 >0$ and hence  convergence can be ensured if $c_1>\frac{b_1^2\delta_0}{(1+\delta_0)^2}$ (With $\eta=\frac{b_1}{c_1(1+\delta_0)}$) for any positive $\delta_0$



Thus from Lemma \ref{recurse} we have that given any $\epsilon >0, \delta \in (0,1)$, $X_{\rm T} \leq \epsilon^2 \cdot \delta$ for, 

\[ T= 1 + \frac{\log \frac{X_1}{\epsilon^2 \delta }}{\log \frac {1}{\alpha} }\text{ with }\alpha=\Big (1 - 2 \eta \lambda_1  + \frac{\eta^2}{b} \cdot (a_4 + a_2^2(b-1) ) \Big ),\eta=\frac{2b\lambda_1}{(a_4 + a_2^2(b-1))(1+\delta_0)} \] 

for a suitable $\delta_0>0$ as mentioned above.

\paragraph{Case 2 : $\theta_* > 0$}
Taking total expectations on both sides of equation \ref{DBothTerms:b} and invoking the RHS of equations \ref{term2:fourth} and \ref{term1:first} we have, 

\begin{align}
    \nonumber X_{t+1} &\leq  \Big (1 - 2 \eta \lambda_1(\theta_*)  + \frac{\eta}{K}  + \frac{\eta^2}{b} \cdot ((1+a_4) + (1+a_2^2)(b-1) ) \Big )X_t\\
    &+   K\theta_* ^2 \cdot \eta \cdot \beta_1^2 + \theta_*^2 \cdot \frac{\eta^2}{b} \cdot \Big ( \beta_3^2 + (\beta_2 \cdot a_1)^2 \cdot (b-1) + (\beta_2 + (b-1)\cdot \beta_1^2) \Big )
\end{align}

Now we can invoke Lemma \ref{recurse2} on the above recursion with the following identifications for the constants therein,  
\[ b_1=2\lambda_1(\theta_*)-\frac{1}{K}, c_1=\frac{1+a_4+(1+a_2^2)(b-1) }{b} \]

\[ c_3=K_1\theta_*^2\beta_1^2, c_2=\frac{\theta_*^2}{\beta_1}\Big ( \beta_3^2 + (\beta_2 \cdot a_1)^2 \cdot (b-1) + (\beta_2 + (b-1)\cdot \beta_1^2) \Big ) \]

Note that since $K$ is so chosen that $2 \lambda_1(\theta_*)>\frac{1}{K}$, we have  $b_1 >0$ and hence the conditions of Lemma \ref{recurse2} 






Hence the smallest value of $X_t$ (say $\epsilon^2 \cdot \delta$ for some $\epsilon >0$ and $\delta \in (0,1)$) that the Lemma \ref{recurse2} guarantees to be attained, say at $X_{\rm T}$ is $\frac{c_3}{b_1}=\frac{K\theta_*^2\beta_1^2}{(2 \lambda_1(\theta)-1/K)}$ for 
$${\rm T} = 
{\large \mathcal{O}}\Bigg(\log \Bigg[~ 
\frac{X_1}{\epsilon^2 \delta  - \Big(\frac{\frac{c_2}{c_1}+\gamma \cdot \frac{c_3}{b_1}}{\gamma - 1}\Big)}~\Bigg]\Bigg) $$ 
when we choose $\eta = \frac{b_1}{\gamma c_1}$ for some $\gamma > \max \left ( \frac{b_1^2}{c_1},\frac{\epsilon^2 \delta + \frac{c_2}{c_1}}{\epsilon^2 \delta    - \frac{c_3}{b_1} } \right).$ Now we can invoke Markov inequality to get what we  set out to prove, 
\[ \mathbb{P} \Big [  \norm{\w_{\rm T} - \w_*}^2 \Big ] \leq \epsilon^2 \Big ] \geq 1 - \delta .\]
\qed 
\end{proofof}






\clearpage 
\section{Estimates for Two Recursions}\label{app:recurse}
\begin{lemma}\label{recurse} 
 Given constants $\eta', b,c_1, c_2 >0$ suppose one has a sequence of real numbers $\Delta_1 = C,\Delta_2,..$ s.t, 

\[ \Delta_{t+1} \leq (1-\eta' b_1 + \eta'^2 c_1)\Delta_t + \eta'^2 c_2  \]

Given any $\epsilon' >0$ in the following two cases we have, $\Delta_{\rm T} \leq \epsilon'^2$ 

\begin{itemize}
    \item If $c_2 =0, C >0$ and for some $\delta_0 >0$ we have, $c_1 > b_1^2 \frac{\delta_0}{(1+\delta_0)^2}$,\\
    $\eta'= \frac {b}{(1+\delta_0))c_1}$ and ${\rm T} = O \Big ( \log \frac{C}{\epsilon'^2 } \Big )$
    
    
    \item If $0 < c_2 \leq c_1,  \epsilon'^2  \leq C, \frac{b^2}{c_1} \leq \Big ( \sqrt{\epsilon'}  + \frac{1}{\sqrt{\epsilon'}}  \Big )^2$,\\
    $\eta' = \frac{b}{c_1}\cdot\frac{\epsilon'^2}{(1+\epsilon'^2)}$ and 
${\rm T}  = O\Bigg(\frac{\log{\bigg(\frac{\epsilon'^2(c_1-c_2)}{Cc_1-c_2\epsilon'^2}}\bigg)}{\log{\bigg(1-\frac{b^2}{c_1}\cdot\frac{\epsilon'^2}{(1+\epsilon'^2)^2}\bigg)}}\Bigg)$
    . 
\end{itemize}
\end{lemma}

\begin{proofof}{Lemma \ref{recurse}}
Suppose we define $\alpha = 1 - \eta' b + \eta'^2 c_1$ and $\beta = \eta'^2 c_2$. Then we have by unrolling the recursion, 
\begin{align*}
\Delta_t &\leq \alpha \Delta_{t-1} + \beta \leq \alpha (\alpha \Delta_{t-1} + \beta ) + \beta \leq  ...\leq \alpha^{t-1}\Delta_1 + \beta \frac{1-\alpha^{t-1}}{1-\alpha}.
\end{align*} 

We recall that $\Delta_1 = C$ to realize that our lemma gets proven if we can find ${\rm T}$ s.t, 
\[\alpha^{{\rm T}-1} C + \beta \frac{1-\alpha^{{\rm T}-1}}{1-\alpha} = {\epsilon'}^2 \] 


Thus we need to solve the following for ${\rm T}$ s.t,  $\alpha ^{{\rm T}-1} = \frac{\epsilon'^2  (1-\alpha) - \beta}{C(1-\alpha) - \beta }$

{\bf Case 1 : $\beta = 0$}
In this case we see that if $\eta >0$ is s.t $\alpha \in (0,1)$ then, 
\[ \alpha ^{{\rm T}-1} = \frac{\epsilon'^2 }{C} \implies {\rm T} = 1 + \frac{\log \frac{C}{\epsilon'^2 }}{\log \frac {1}{\alpha} } \]

But $\alpha = \eta'^2 c_1 - \eta' b + 1 = \Big ( \eta' \sqrt{c_1} - \frac{b}{2\sqrt{c_1}} \Big )^2 + \Big (  1 - \frac{b^2}{4c_1} \Big )$ Thus $\alpha \in (0,1)$ is easily ensured by choosing $\eta' =\frac{b_1}{(1+\delta_0)c_1}$ for some $\delta_0>0$ and  $c_1> b_1^2 \frac{\delta_0}{(1+\delta_0)^2}$

This gives us the first part of the theorem.

{\bf Case $2$ : $\beta >0$}

This time we are solving, 

\begin{align}\label{betanonz}
\alpha^{{\rm T}-1} = \frac{\epsilon'^2  (1-\alpha)-\beta}{C(1-\alpha)-\beta}
\end{align} 

Towards showing convergence, we want to set $\eta'$ such that $ \alpha^{t-1} \in (0,1)$ for all $t$. Since $\epsilon'^2  < C$, it is sufficient to require, 

\begin{align*}
    \beta < \epsilon'^2(1-\alpha) &\implies \alpha < 1 - \frac{\beta}{\epsilon'^2}
    \Leftrightarrow 1-\frac{b^2}{4c_1} + \Big ( \eta' \sqrt{c_1} - \frac{b}{2\sqrt{c_1}} \Big )^2 \leq 1 - \frac{\beta}{\epsilon'^2}\\ 
    &\Leftrightarrow \frac{\eta'^2c_2}{\epsilon'^2} \leq \frac{b^2}{4c_1} - \Big ( \eta' \sqrt{c_1} - \frac{b}{2\sqrt{c_1}} \Big )^2 \Leftrightarrow \frac{c_2}{\epsilon'^2} \leq \frac{b^2}{4c_1\eta'^2} - \Big ( \sqrt{c_1} - \frac{b}{2\sqrt{c_1}\eta'} \Big )^2 
\end{align*}

Set $\eta' = \frac{b}{\gamma c_1}$ for some constant $\gamma>0$ to be chosen such that,
\begin{align*}
    \frac{c_2}{\epsilon'^2} \leq \frac{b^2}{4c_1 \cdot\frac{b^2}{\gamma^2c_1^2}} - \Big ( \sqrt{c_1} - \frac{b}{2\sqrt{c_1}\cdot\frac{b}{\gamma c_1}} \Big )^2 \implies \frac{c_2}{\epsilon'^2} \leq c_1\frac{\gamma^2}{4} - c_1\cdot\Big (\frac{\gamma}{2} - 1 \Big )^2 \implies  c_2 \leq \epsilon'^2 \cdot c_1 (\gamma-1)
\end{align*}

Since $c_2 \leq c_1$ we can choose, $\gamma = 1+\frac{1}{\epsilon'^2}$ and we have  $\alpha^{t-1} < 1$.  
Also note that, 
\begin{align*}
\alpha & = 1+ \eta'^2c_1 - \eta'b = 1+ \frac{b^2}{\gamma^2c_1^2} - \frac{b^2}{\gamma c_1}  = 1 - \frac{b^2}{c_1}\cdot\big(\frac{1}{\gamma}-\frac{1}{\gamma^2}\big). \\
& = 1 - \frac{b^2}{c_1}\cdot \frac{\epsilon'^2}{(1+\epsilon'^2)^2} = 1 - \frac{b^2}{c_1} \cdot \frac{1}{\Big ( \epsilon'  + \frac{1}{\epsilon' }  \Big )^2} 
\end{align*}

And here we recall that the condition that the lemma specifies on the ratio $\frac{b^2}{c_1}$ which ensures that the above equation leads to $\alpha >0$ 

Now in this case we get the given bound on ${\rm T}$ in the lemma by solving equation \ref{betanonz}. To see this, note that,
\begin{align*}
\alpha = 1-\frac{b^2}{c_1}\cdot\frac{\epsilon'^2}{(1+\epsilon'^2)^2}
\text{ and }
\beta = \eta'^2 c_2 = \frac{b^2}{\gamma^2 c_1} \cdot c_2 = \frac{b^2c_2}{c_1}\cdot \frac{(\epsilon'^2)^2}{(1+\epsilon'^2)^2}.
\end{align*}

Plugging the above into equation \ref{betanonz} we get, 
$\alpha^{{\rm T}-1} = \frac{\epsilon'^2\Delta_(c_1-c_2)}{Cc_1-c_2\epsilon'^2} 
\implies {\rm T} = 1 + \frac{\log{\bigg(\frac{\epsilon'^2(c_1-c_2)}{Cc_1-c_2\epsilon'^2}}\bigg)}{\log{\bigg(1-\frac{b^2}{c_1}\cdot\frac{\epsilon'^2}{(1+\epsilon'^2)^2}\bigg)}}$.
\qed
\end{proofof}

\begin{lemma}\label{recurse2} 
Suppose we have a sequence of real numbers $\Delta_1, \Delta_2, \ldots$ s.t 
\[\Delta_{t+1} \leq (1-\eta' b_1 + \eta'^2 c_1) \Delta_t + \eta'^2c_2 + \eta'c_3\]
for some fixed parameters $b_1,c_1, c_2, c_3 >0$ s.t $\Delta_1 > \frac{c_3}{b_1}$ and free parameter $\eta' > 0$. Then for,

\[ ~\epsilon'^2 \in \Big ( \frac{c_3}{b_1} ,\Delta_1 \Big ),\quad ~\eta' = \frac{b_1}{\gamma c_1}, \quad \gamma > \max \Bigg \{ \frac{b_1^2}{c_1},  \Bigg ( \frac{\epsilon'^2 + \frac{c_2}{c_1}}{\epsilon'^2   - \frac{c_3}{b_1}} \Bigg ) \Bigg \}  > 1\]

it follows that $\Delta_{\rm T} \leq \epsilon'^2$ for, 
$${\rm T} = 
{\large \mathcal{O}}\Bigg(\log \Bigg[~ 
\frac{\Delta_1}{{\epsilon'}^2 - \Big(\frac{\frac{c_2}{c_1}+\gamma \cdot \frac{c_3}{b_1}}{\gamma - 1}\Big)}~\Bigg]\Bigg) $$ 
\end{lemma} 

\begin{proofof}{Lemma \ref{recurse2}}
Let us define $\alpha = 1-\eta' b_1 + \eta'^2 c_1$ and $\beta = \eta'^2c_2 + \eta'c_3$.
Then by unrolling the recursion we get, 
\[ \Delta_t \leq \alpha \Delta_{t-1} + \beta \leq \alpha (\alpha \Delta_{t-2} + \beta ) + \beta \leq  ...\leq \alpha^{t-1}\Delta_1 + \beta (1+\alpha+\ldots+\alpha^{t-2}). \]

Now suppose that the following are true for $\epsilon'$ as given and for $\alpha ~\& ~\beta$ (evaluated for the range of $\eta'$s as specified in the theorem), 
\begin{itemize}
    \item[ ] {\bf Claim 1 :} $\alpha \in (0,1)$
    \item[ ] {\bf Claim 2 :} $0 < \epsilon'^2(1-\alpha) - \beta$
\end{itemize}

We will soon show that the above claims are true. Now if ${\rm T}$ is s.t we have, \[ \alpha^{{\rm T}-1}\Delta_1 + \beta (1+\alpha+\ldots+\alpha^{{\rm T}-2}) 
= \alpha^{{\rm T}-1}\Delta_1 + \beta \cdot \frac{1 - \alpha^{\rm T-1}}{1-\alpha}  = \epsilon'^2 \] then $\alpha^{{\rm T} -1} = \frac{\epsilon'^2(1-\alpha) - \beta}{\Delta_1(1-\alpha) - \beta}$. 
Note that {\bf Claim 2} along with with the assumption that $\epsilon'^2 < \Delta_1$ ensures that the numerator and the denominator of the fraction in the RHS are both positive. Thus we can solve for ${\rm T}$ as follows,  

\begin{align}\label{eq:recursionbound}
\implies \nonumber ({\rm T}-1) \log\left(\frac{1}{\alpha}\right) &= \log \bigg[\frac{\Delta_1(1-\alpha)-\beta}{\epsilon'^2  (1-\alpha)-\beta}\bigg] \implies {\rm T}  = {\large \mathcal{O}}\Bigg(\log \Bigg[~ 
\frac{\Delta_1}{\epsilon'^2 - \Big(\frac{\frac{c_2}{c_1}+\gamma \cdot \frac{c_3}{b}}{\gamma - 1}\Big)}~\Bigg]\Bigg)
\end{align} 
In the second equality above we have estimated the expression for ${\rm T}$ after substituting $\eta' = \frac{b_1}{\gamma c_1}$ in the expressions for $\alpha$ and $\beta$.
\qed 
\end{proofof}
~\\

\begin{proofof}{claim 1 : $\alpha \in (0,1)$}

We recall that we have set $\eta' = \frac{b_1}{\gamma c_1}$. This implies that,
$\alpha  = 1 - \frac{b_1^2}{c_1}\cdot \Big ( \frac {1}{\gamma} - \frac{1}{\gamma^2} \Big )$. Hence $\alpha > 0$ is ensured by the assumption that $\gamma > \frac{b_1^2}{c_1}$. And $\alpha < 1$ is ensured by the assumption that $\gamma >1$ 

\qed 
\end{proofof}

\begin{proofof}{claim 2 : $0 < \epsilon'^2(1-\alpha) - \beta$}

We note the following, 
\begin{align*}
    - \frac{1}{\epsilon'^2} \cdot \left(\epsilon'^{2}(1-\alpha) -\beta\right) & =\alpha - \Big (1 - \frac{\beta}{\epsilon'^2} \Big ) \\
    & =  1-\frac{b_1^2}{4c_1} + \Big ( \eta' \sqrt{c_1} - \frac{b_1}{2\sqrt{c_1}} \Big )^2  - \Big ( 1 - \frac{\beta}{\epsilon'^2}  \Big )\\
    &=\frac{\eta'^2c_2 + \eta'c_3}{\epsilon'^2} + \Big ( \eta' \sqrt{c_1} - \frac{b_1}{2\sqrt{c_1}} \Big )^2 - \frac{b_1^2}{4c_1}\\
    &= \frac{\left(\eta'\sqrt{c_2} + \frac{c_3}{2\sqrt{c_2}}\right)^2 -  \frac{c_3^2}{4c_2}}{\epsilon'^2} + \Big ( \eta' \sqrt{c_1} - \frac{b_1}{2\sqrt{c_1}} \Big )^2 - \frac{b_1^2}{4c_1}\\
    &= \eta'^2 \Bigg (  \frac{1}{\epsilon'^2} \cdot \left(\sqrt{c_2} + \frac{c_3}{2\eta'\sqrt{ c_2}}\right)^2 + \Big (\sqrt{c_1} - \frac{b_1}{2\eta ' \sqrt{c_1}} \Big )^2 - \frac{1}{\eta'^2} \Bigg[  \frac{b_1^2}{4c_1} +  \frac{1}{\epsilon'^2}\left( \frac{c_3^2}{4c_2}\right) \Bigg] \Bigg ) 
\end{align*}

Now we substitute $\eta' = \frac{b_1}{\gamma c_1}$ for the quantities in the expressions inside the parantheses to get, 

\begin{align*}
     - \frac{1}{\epsilon'^2} \cdot \left(\epsilon'^{2}(1-\alpha) -\beta\right) 
    & = \alpha - \Big (1 - \frac{\beta}{\epsilon'^2} \Big )\\
    &= \eta'^2 \Bigg ( \frac{1}{\epsilon'^2 } \cdot \left(\sqrt{c_2} + \frac{\gamma c_1c_3}{2b_1\sqrt{ c_2}}\right)^2 +  c_1\cdot\Big (\frac{\gamma}{2} - 1 \Big )^2 - c_1\frac{\gamma^2}{4} - \frac{1}{\epsilon'^2}\cdot \frac{\gamma^2c_1^2c_3^2}{4b_1^2c_2}  \Bigg )\\
    &= \eta'^2 \Bigg ( \frac{1}{\epsilon'^2} \cdot \left(\sqrt{c_2} + \frac{\gamma c_1c_3}{2b_1\sqrt{ c_2}}\right)^2 + c_1(1-\gamma) - \frac{1}{\epsilon'^2}\cdot \frac{\gamma^2c_1^2c_3^2}{4b_1^2c_2} \Bigg )\\
    &= \frac{\eta'^2}{\epsilon'^2} \Bigg (   c_2 + \frac{\gamma c_1c_3}{b_1} - \epsilon'^2  c_1(\gamma -1) \Bigg )\\
    &= \frac{\eta'^2c_1}{\epsilon'^2} \Bigg (   (\epsilon'^2   + \frac{c_2}{c_1})  - \gamma \cdot \left (\epsilon'^2 - \frac{c_3}{b_1} \right )  \Bigg ) 
\end{align*}    

Therefore, $-\frac{1}{\epsilon'^2}\left(\epsilon'^{2}(1-\alpha) - \beta\right) < 0$ since by assumption $\epsilon'^2   > \frac{c_3}{b_1},~\text{ and }~ \gamma > \left(\epsilon'^2 + \frac{c_2}{c_1}\right)/\left(\epsilon'^2 - \frac{c_3}{b_1}\right)$.

\qed
\end{proofof}

\end{document}